\documentclass[journal]{IEEEtran}

\usepackage{algorithm}
\usepackage{algpseudocode}

\usepackage[english]{babel}
\usepackage[utf8x]{inputenc}
\usepackage{indentfirst}
\usepackage{amsmath}
\usepackage{graphicx}
\usepackage[colorinlistoftodos]{todonotes}
\usepackage{tikz}
\usetikzlibrary{arrows.meta, positioning}
\usepackage{listings}
\usepackage{float}
\usepackage{booktabs} % For better table formatting
\usepackage[numbers]{natbib}
\usepackage{amssymb}
\usepackage{booktabs}
\usepackage{multirow}
\usepackage{float}
\usepackage{tabularx}

\usepackage{titlesec}
\ifCLASSINFOpdf
\else
\fi
\begin{document}
\title{Timing the Match: A Deep Reinforcement Learning Approach for Ride-Hailing and Ride-Pooling Services}

\author{Yiman Bao,%~~~
        ~Jie Gao$^*$,%
        ~Jinke He,%
        ~Frans A. Oliehoek,%
        ~Oded Cats% <-this % stops a space
\thanks{Yiman Bao, Delft University of Technology, Delft, The Netherlands, the department of Transport \& Planning  (e-mail: baoyiman5@gmail.com).}%
\thanks{Jie Gao, corresponding author, Delft University of Technology, Delft, The Netherlands, the department of Transport \& Planning  (e-mail: J.Gao-1@tudelft.nl).}%
\thanks{Jinke He, Delft University of Technology, Delft, The Netherlands, the department of Intelligent Systems (e-mail: J.He-4@tudelft.nl).}%
\thanks{Frans A. Oliehoek, Delft University of Technology, Delft, The Netherlands, the department of Intelligent Systems (e-mail: F.A.Oliehoek@tudelft.nl).}%
\thanks{Oded Cats, Delft University of Technology, Delft, The Netherlands, the department of Transport \& Planning (e-mail: o.cats@tudelft.nl).}%
}

\maketitle

\begin{abstract}
Efficient timing in ride-matching is crucial for improving the performance of ride-hailing and ride-pooling services, as it determines the number of drivers and passengers considered in each matching process. Traditional batched matching methods often use fixed time intervals to accumulate ride requests before assigning matches. While this approach increases the number of available drivers and passengers for matching, it fails to adapt to real-time supply-demand fluctuations, often leading to longer passenger wait times and driver idle periods.
To address this limitation, we propose an adaptive ride-matching strategy using deep reinforcement learning (RL) to dynamically determine when to perform matches based on real-time system conditions. Unlike fixed-interval approaches, our method continuously evaluates system states and executes matching at moments that minimize total passenger wait time. Additionally, we incorporate a potential-based reward shaping (PBRS) mechanism to mitigate sparse rewards, accelerating RL training and improving decision quality. 
Extensive empirical evaluations using a realistic simulator trained on real-world data demonstrate that our approach outperforms fixed-interval matching strategies, significantly reducing passenger waiting times and detour delays, thereby enhancing the overall efficiency of ride-hailing and ride-pooling systems.
%An efficient simulator is designed to model the spatiotemporal matching process in ride-hailing and ride-pooling services based on real world data. Through experiments, the reinforcement learning-based matching strategy is compared with fixed-interval and real-time matching strategies. The results demonstrate that the reinforcement learning-based strategy outperforms the others in key metrics such as total waiting time and detour delay. 
\end{abstract}

\begin{IEEEkeywords}
Deep reinforcement learning, ride-hailing, ride-pooling, matching optimization, matching time, proximal policy optimization.
\end{IEEEkeywords}

\section{Introduction}
\IEEEPARstart{R}{i}de-hailing services, such as Uber\footnote{https://www.uber.com/nl/en/} and Lyft \footnote{https://www.lyft.com/}, have become integral to urban transportation, offering passengers flexible, on-demand mobility while optimizing vehicle dispatch to alleviate congestion and reduce emissions \cite{Brown2020Hailing, Wei2021Transit}. As an extension of ride-hailing, ride-pooling enables multiple passengers with similar routes to share a single vehicle, improving vehicle utilization and lowering travel costs, contributing further to sustainable urban mobility \cite{Shaheen2018Shared, Ke2020On}. However, as these services expand, ensuring efficient matching in the face of dynamic supply and demand remains a fundamental challenge. 

To improve matching efficiency, ride-hailing platforms typically employ two primary strategies. One approach is first dispatch, where requests are assigned as they arrive based on available drivers. While this minimizes response time, it often leads to suboptimal matches, as better driver-passenger pairings could emerge if requests were aggregated over time. To address this, platforms such as Uber adopt batched matching, where ride requests and available drivers are accumulated over a fixed time interval before matches are assigned \cite{uber_batched}. By increasing the pool of potential matches, this method improves assignment quality and enhances ride-pooling efficiency. However, fixed-interval batching does not account for real-time fluctuations in supply and demand. Performing matches at predetermined intervals may not coincide with the optimal matching moment, where additional waiting no longer improves the quality of matches. As a result, fixed batching strategies can lead to unnecessary passenger delays or missed opportunities for more efficient assignments.

The dynamic and stochastic nature of ride-hailing environments makes it challenging to determine optimal matching timing using fixed-rule strategies. Instead, an adaptive approach that learns to identify these moments based on real-time system conditions is needed. Reinforcement Learning (RL), a framework for sequential decision-making under uncertainty, is well-suited for this problem. By modeling ride-matching as a Markov Decision Process (MDP) \cite{puterman2014markov}, RL can continuously evaluate system states and dynamically adjust matching timing to balance efficiency and responsiveness. In this work, we propose an RL-based adaptive matching strategy that determines when to execute matches based on real-time system conditions. Unlike fixed-interval batching, our approach continuously monitors supply-demand variations and optimizes timing to minimize passenger wait times. To address sparse reward challenges and accelerate learning, we introduce a potential-based reward shaping (PBRS) mechanism that improves RL training efficiency. Furthermore, we develop a realistic ride-matching simulator trained on real-world data, enabling comprehensive evaluation of our approach against existing first dispatch and batched matching strategies.
The main contributions of this work are as follows:%Compared to rule-based heuristics or conventional optimization methods, RL offers greater adaptability, as it can learn from both historical and real-time data, refining its strategy over time.

\begin{enumerate}
    \item We formulate the matching process of ride-hailing and ride-pooling as a Partially Observed MDP (POMDP). At each time step, the match-maker determines whether the current moment is optimal for matching and decides whether to execute the matching. Once a match is confirmed, the match-maker performs the optimal matching between idle drivers and unserved passengers in the accumulated pool.
    \item We design a RL framework based on the formulated POMDP, including actions, states, and rewards. To address the issue of sparse rewards, we introduce Potential Based Reward Shaping (PBRS), which proofs that rewards are appropriately allocated to each action.
    \item We develop an efficient and realistic simulator to model the matching process of ride-hailing and ride-pooling. This simulator, based on real-world data, includes functionalities such as data generation and spatial matching. We conduct a series of training and validation experiments using this simulator.
    \item We employ Proximal Policy Optimization (PPO) as the RL method and investigate the impact of applying PBRS on the training outcomes. Additionally, we explore how the decision-making effectiveness of the match-maker evolves across different training phases.
    \item We compare the strategies trained by RL with the first dispatch strategy and batched matching strategies at different time intervals. Our approach demonstrates superior performance compared to the latter two. Furthermore, we analyze the adaptability of our approach, revealing that it flexibly makes different action decisions in response to varying supply and demand conditions.
\end{enumerate}

The remainder of the paper is structured as follows: Section 2 reviews the related literature and summarizes the research gap. Section 3 introduces the proposed approach for optimizing the timing for batched matching. Section 4 introduces the designed simulator. Section 5 conducts extensive numerical experiments on a real-world data. Finally, conclusions are given in Section 6.
%Through these contributions, this study offers a flexible and adaptive optimization framework that reduces passenger waiting time, and provides a foundation for future intelligent mobility services in urban environments.

\section{Literature review}
\subsection{Matching in Ride-hailing Systems}
In ride-hailing systems, matching strategies are designed to efficiently allocate available drivers to passenger requests, ensuring timely pickups and maximizing system-wide efficiency. The matching process is typically formulated as an optimization problem, where assignments are made to maximize an objective function such as minimizing passenger wait times or maximizing driver utilization. Yan et al. \cite{yan2020dynamic} modeled this problem as a bipartite graph, where nodes represent drivers and passengers, and edges indicate potential matches, weighted by their quality. The optimization objective is to maximize the total weighted matchings. In some studies, the ride-hailing matching process is also modeled as a game among passengers, drivers, and the platform, where each party seeks a matching outcome that maximizes its own benefit. Li et al. \cite{li2024allocation} formulate this problem as a Stackelberg game, in which the ride-hailing platform acts as the leader, setting prices to ensure that none of the three parties can achieve a more profitable allocation, thereby determining an outcome that maximizes their respective profits.

Beyond static optimization formulations, dynamic modeling approaches have been proposed to account for real-time decision-making in ride-hailing systems. Beojone et al. \cite{BEOJONE2023104375} developed a mixed continuous-discrete time Markov Chain model to predict driver earnings and mobility patterns, enabling more informed dispatching. Similarly, Zhang et al. \cite{ZHANG2023102819} formulated idle driver routing as a Markov Decision Process (MDP), improving vehicle utilization by optimizing movement strategies before new requests arrive. These studies highlight the importance of incorporating dynamic system states into ride-matching models, providing valuable insights for real-time decision-making.

Another important factor influencing ride-hailing efficiency is driver heterogeneity, which affects individual responses to ride requests. Do et al. \cite{su11205615} analyzed factors leading to driver request rejections, while Liu et al. \cite{LIU2024104466} applied inverse reinforcement learning to detect anomalous driver behavior and improve matching success rates. Additionally, Shi et al. \cite{SHI2023169} observed that part-time drivers exhibit lower patience levels than full-time drivers, making them more sensitive to wait times. To address this, they proposed a priority-based matching policy (PMP) that prioritizes part-time drivers to reduce abandonment rates. With the growing adoption of electric vehicles (EVs), ride-hailing matching must consider EV-specific constraints such as range limits and charging delays. Li et al. \cite{li2024bm} proposed a stochastic optimization framework that integrates vehicle repositioning and matching. The framework proactively guides idle EVs based on demand forecasts and optimizes matching to minimize passenger wait times and charging-related costs. Furthermore, some studies have explored the integration of driver rebalancing with matching strategies. Guo et al. \cite{GUO2021161} introduced the Matching-Integrated Vehicle Rebalancing (MIVR) model, which simultaneously considers passenger-driver assignments and vehicle repositioning to enhance fleet efficiency.

In addition to spatial assignment strategies, matching quality is influenced by both the matching radius and timing. Yang et al. \cite{YANG202084} identified these two parameters as key determinants of system performance. A shorter matching radius reduces pick-up distances but may lower the overall matching rate, whereas a longer matching interval allows for the accumulation of more unserved passengers and idle drivers, potentially improving assignment quality at the cost of increased passenger wait times. To optimize these trade-offs, various approaches have been proposed. Chen et al. \cite{chen2025dynamic} developed a Deep Learning-based Matching Radius Decision (DL-MRD) model that predicts key system performance metrics across different matching radii, enabling the ride-hailing platform to select an optimal radius based on real-time supply and demand conditions. To account for the temporal impact on matching outcomes, Shi et al. \cite{shi2024vehicle} consider how current matching and dispatch decisions influence future ride-hailing supply and demand, leading to variations in vehicle value across different regions and time periods. They introduce a vehicle value function to quantify the spatiotemporal value of vehicles in each region and incorporate it into the design of order-matching and idle vehicle dispatch algorithms.

\subsection{Matching in Ride-pooling Systems}
Ride-pooling introduces additional complexity compared to ride-hailing, as it requires not only the assignment of drivers to passengers but also the coordination of passengers with similar routes while optimizing pick-up and drop-off sequences to minimize detours. Agatz et al. \cite{AGATZ2012295} categorized ride-sharing into several types, including single-driver, single-rider; single-driver, multiple-riders; multiple-drivers, single-rider; and multiple-drivers, multiple-riders. Among these, ride-hailing corresponds to the single-driver, single-rider case, whereas ride-pooling aligns with the single-driver, multiple-rider model, introducing additional constraints related to route compatibility, shared occupancy optimization, and dynamic reassignments.

A variety of optimization-based approaches have been developed to address ride-pooling matching. Alonso-Mora et al. \cite{Alonso-Mora} proposed an anytime optimal algorithm that balances fleet size, vehicle capacity, passenger wait times, trip detours, and operational costs. Their method begins with a greedy allocation and iteratively refines the solution through constrained optimization, ensuring computational scalability while converging to an optimal allocation. Li et al. \cite{li2021optimizing} focused on scenarios where each driver serves at most two passengers, introducing a rolling horizon approach with financial incentives to encourage ride-pooling adoption. Their model, formulated as a multi-stage integer programming problem, seeks to maximize system-wide profitability. Chen et al. \cite{chen2024network} modeled ride-pooling services using a layered OD graph, decomposing the matching process into three subproblems: vehicle dispatching, order pooling sequence, and routing optimization. A minimum-cost network flow model was employed to optimize the order matching sequence.

Given the computational challenges of real-time ride-pooling matching, various studies have explored ways to improve algorithmic efficiency. Simonetto et al. \cite{SIMONETTO2019208} formulated the ride-pooling problem as a linear assignment problem between vehicles and passenger requests, employing a federated optimization framework to enhance computational speed. Meshkani et al. \cite{MESHKANI2023100106} introduced a decentralized ride-pooling model based on vehicle-to-infrastructure (V2I) and infrastructure-to-infrastructure (I2I) communication, achieving a 25.53-fold improvement in computational efficiency over conventional methods.

Some studies have sought to jointly optimize ride-hailing and ride-pooling operations by developing unified matching models. Qin et al. \cite{QIN2021103287} proposed a multi-objective integer linear programming framework that simultaneously considers three service modes: ride-pooling (RP), non-ride-pooling (NP), and a "bundled" hybrid model. Their two-stage Kuhn-Munkres (2-KM) algorithm iteratively refines passenger-vehicle assignments, ensuring an adaptive balance between shared and private rides. Similarly, Zhou et al. \cite{ZHOU2023104326} introduced a decision-making framework for ride-hailing platforms that optimizes passenger-vehicle matching while considering passengers’ monetary and travel experience trade-offs.

Matching timing plays a crucial role in determining the effectiveness of ride-pooling. A longer matching time window allows for accumulating a larger set of potential passengers with compatible routes, reducing overall detour distances and improving efficiency. However, longer wait times may negatively impact user experience. The challenge, therefore, lies in dynamically determining the optimal moment to execute matching decisions rather than relying on predefined fixed-interval batching methods. Guo et al. \cite{GUO2021810} introduced a real-time ride-pooling framework with dynamic time windows and expectation-based migration, where ride requests can be strategically deferred to future time windows based on historical demand patterns. Their multi-strategy graph search heuristic improves scalability, enabling the system to handle large-scale ride-pooling scenarios. Zhao et al. \cite{zhaooptimizing} proposed an embedding-based online matching framework that adapts to network structure changes. This approach constructs a dynamic heterogeneous ride-pooling network incorporating various node attributes and driver-passenger connections. By continuously updating representations to capture network evolution, it efficiently identifies and ranks candidate passengers for real-time matching. Furthermore, Yang et al. \cite{yang2024prediction} proposed a proactive vehicle dispatch strategy that forecasts future supply and demand using a Poisson distribution. The strategy estimates potential distance savings from future orders and solves a bipartite matching problem to assign passengers to partially occupied or idle vehicles or defer them to the next matching round. However, most traditional operations research approaches rely on static demand forecasts and do not incorporate real-time adaptive decision-making, limiting their effectiveness in highly dynamic environments.

Given the high uncertainty in real-time ride-pooling demand, existing approaches that rely on predefined matching intervals or offline demand estimation lack the flexibility to adapt to evolving conditions. This study addresses these limitations by developing a reinforcement learning (RL)-based approach that optimizes the timing of ride-pooling matches, enabling real-time adjustments to supply-demand variations. By leveraging RL, the proposed method learns when to execute ride-matching decisions based on system conditions rather than relying on fixed rules or heuristic estimations. This allows the system to continuously adapt to changing demand patterns, improving efficiency while maintaining acceptable passenger wait times.

\subsection{RL for Ride-hailing and Ride-pooling Systems}
In real-world ride-hailing and ride-pooling systems, supply and demand conditions are highly dynamic and continuously evolving, posing challenges for effective ride-matching. Reinforcement learning (RL) has emerged as a promising approach for addressing these complexities due to its adaptive decision-making capabilities. By continuously learning from changing market conditions, RL can optimize decision policies while adapting to uncertainties in real-time demand fluctuations. Additionally, its scalability and transferability enable rapid deployment across different urban environments, ensuring sustained operational efficiency.

Several studies have explored RL applications in ride-hailing and ride-pooling to address various optimization challenges. Qiao et al. \cite{QIAO2023110965} proposed a multi-agent RL-based algorithm (ERPM) for intelligent ride-hailing demand prediction. ERPM mitigates convergence issues in traditional RL models, which arise from the high dimensionality of grid-based demand forecasting, by leveraging the Actor–Critic strategy to optimize ride-hailing dispatch decisions. Similarly, Zhang et al. \cite{zhang2024nondbrem} developed NondBREM, an offline deep reinforcement learning framework for large-scale ride-hailing order dispatch. To avoid costly and unsafe interactions with the environment, it learns solely from historical data. The framework incorporates a Nondeterministic Batch-Constrained Q-learning (NondBCQ) module to reduce extrapolation errors and a Random Ensemble Mixture (REM) module to enhance generalization and robustness. For ride-pooling, Hu et al. \cite{hu2025bmg} proposed the Localized Bipartite Match Graph Attention Q-Learning (BMG-Q) framework, specifically designed for ride-pooling order dispatch. It achieves approximately $10\%$ higher cumulative rewards than baseline reinforcement learning models while reducing overestimation bias by over $50\%$ . Additionally, BMG-Q demonstrates robustness to task and fleet size variations, making it an effective, scalable, and resilient framework for ride-pooling operations.

A closely related study by Ke et al. \cite{ke2020learning} applied multi-agent RL to determine whether passengers should delay matching to improve their pairing outcome. Their approach treats each passenger as an independent agent, benefiting from multiple reward signals per time step due to the continuous arrival of ride requests, thereby mitigating sparse reward issues. However, their method optimizes individual waiting decisions rather than system-wide efficiency, potentially leading to locally optimal rather than globally optimal solutions. Additionally, their study does not consider ride-pooling scenarios, limiting its applicability to shared mobility services with more complex passenger-routing constraints. In contrast, this study models the ride-matching system as a single learning agent, optimizing matching timing across both ride-hailing and ride-pooling. Unlike fixed-time matching strategies, which operate on predefined intervals regardless of real-time conditions, our approach dynamically determines when to execute matching based on observed supply-demand fluctuations. Additionally, rather than optimizing individual passenger decisions, as seen in Ke et al. \cite{ke2020learning}, our method seeks a globally optimal matching policy that enhances overall system efficiency. Furthermore, while most RL-based studies focus exclusively on ride-hailing, our framework extends to ride-pooling, capturing the added complexity of shared trips and optimizing both driver-passenger and passenger-passenger assignments. The positioning of our work relative to existing research is summarized in Table \ref{tab:Literature_Review}.

\begin{table}[H] % Use [H] to enforce exact placement
\centering
\caption{Research Comparison}
\label{tab:Literature_Review}
\resizebox{\columnwidth}{!}{%
\begin{tabular}{lccccc}
\toprule
Researchers                                         & Ride-Hailing & Ride-Pooling & Spatial Matching & Temporal Matching & RL \\ \midrule
Yan et al.\cite{yan2020dynamic}         & \checkmark   &              & \checkmark      &                   &    \\
Zhang et al.\cite{ZHANG2023102819}              & \checkmark   &              & \checkmark      &                   &    \\
Yang et al.\cite{YANG202084} & \checkmark            &    & \checkmark      & \checkmark                  &    \\
Shi et al.\cite{shi2024vehicle}            & \checkmark             &    & \checkmark      & \checkmark             &    \\
Agatz et al.\cite{AGATZ2012295}            &              & \checkmark   & \checkmark      &                   &    \\
Alonso-Mora et al.\cite{Alonso-Mora}              &              & \checkmark   & \checkmark      &                   &    \\
Qin et al.\cite{QIN2021103287}     & \checkmark      & \checkmark   & \checkmark      &                   &    \\
Yang et al.\cite{yang2024prediction}           &    & \checkmark   & \checkmark      & \checkmark            &    \\Qiao et al.\cite{QIAO2023110965} & \checkmark & & & & \checkmark \\
Hu et al.\cite{hu2025bmg} & & \checkmark & & & \checkmark \\
 et al.\cite{ke2020learning}           & \checkmark   &        &   \checkmark        & \checkmark        & \checkmark \\
Our Research                                        & \checkmark   & \checkmark   & \checkmark      & \checkmark        & \checkmark \\ \bottomrule
\end{tabular}%
}
\end{table}

\section{Optimizing the timing of batched matching}
This section introduces the methodology developed to optimize the timing for batched matching in ride-hailing and ride-pooling systems. 
First, we provide a general description of the problem.
Second, we model the problem as a sequential decison-making problem under the reinforcement learning (RL) framework.
Lastly, we describe how to solve this optimization problem with the popular Proximal Policy Optimization (PPO) algorithm \cite{schulman2017ppo}.
%The RL framework focuses on designing state and action spaces and formulating a reward function to address the matching decision problem. And the PPO algorithm is employed to train the RL framework to determine the matching timing for ride-hailing and ride-pooling systems.
\subsection{Batched matching}
Batched matching is an optimization problem that assigns drivers to rider requests collected within a batching window. The goal is to maximize or minimize certain objective functions—such as social, economic, or service quality metrics—relative to a set of feasible assignments.
\begin{figure}[h]
    \centering
    \includegraphics[width=\linewidth]{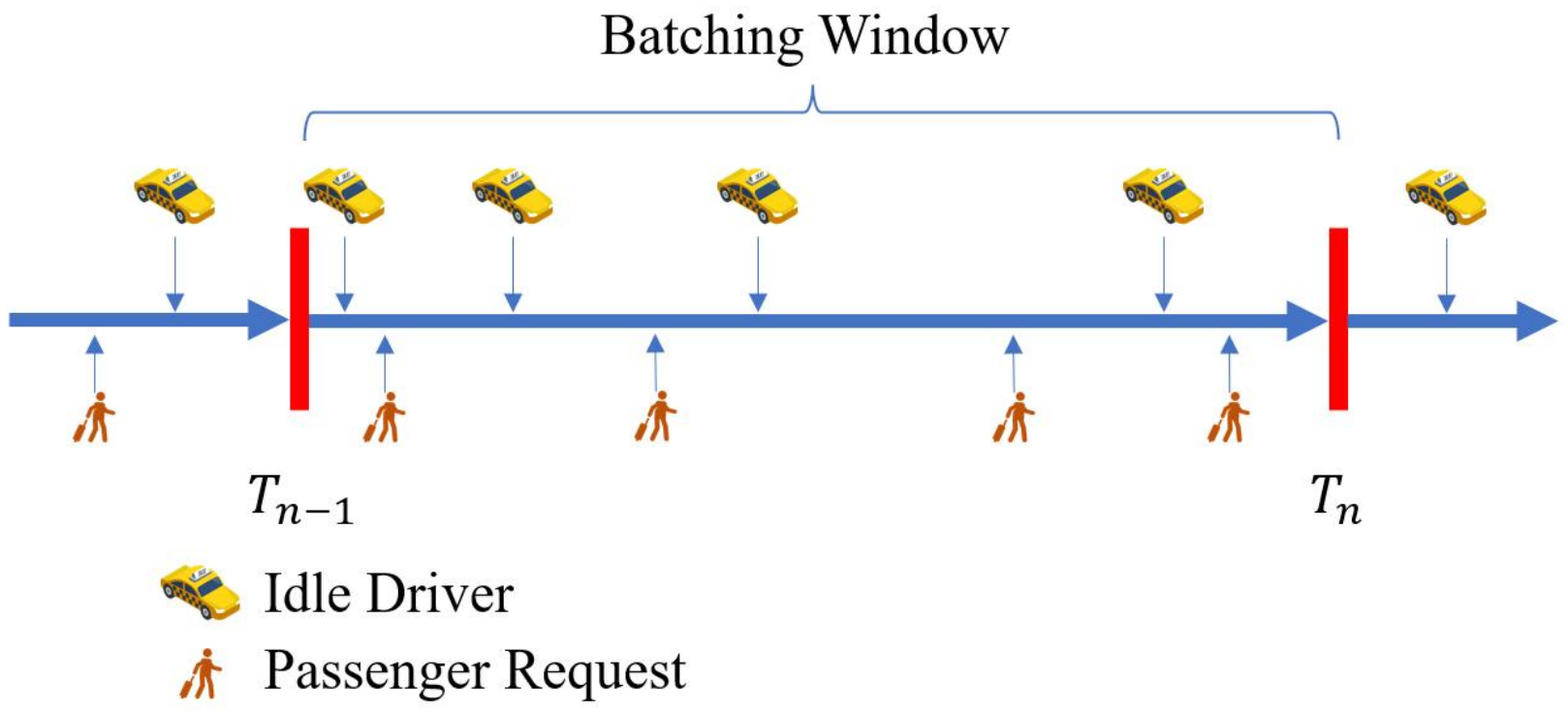} 
    \caption{Comparison of Rewards Before and After PBRS} 
    \label{fig:batch_matching} 
\end{figure}
As illustrated in Figure~\ref{fig:batch_matching}, rider requests are collected during a pre-defined batching window (e.g., $T_n$ in Fig.~\ref{fig:batch_matching}). At the end of this window, a batched matching algorithm computes the assignments and sends the results to both drivers and riders. Any unmatched rider requests from this batch are carried over and resolved in subsequent batching windows.

Formally, we divide the time of a day into a set of time windows $\{T_0, ..., T_n,..\}$ where $T_n = [t_{n-1}, t_n]$ and $t_n - t_{n-1} = \Delta T ~\forall n$. Instead of executing the matching algorithm at every predefined $t_n$, we aim to dynamically determine the optimal timing for conducting the matching. 
That is, find the best $t_n$ throughout the day to execute the matching algorithm. The objective is to minimize the total waiting time in the system by dynamically adjusting the batching window based on system conditions.

\subsection{Deciding when to match as an RL task}
The problem of determining when to match can be naturally formalized as a sequential decision-making problem and thus an RL task.
By formulating the problem as a Markov Decision Process \cite{puterman_markov_1994}, we enable the training of an adaptive strategy for determining the optimal matching moments with RL methods.

%Key components of this framework include the design of action and state spaces tailored to our decision-making problems, along with a customized reward function that leverages Potential-based Reward Shaping (PBRS) \cite{ng1999policy} to mitigate the issue of sparse rewards, which has been shown crucial for learning efficiency.

\subsubsection{Finite-horizon Markov Decision Process}
The process of determining when to match in ride-hailing and ride-pooling services can be formulated as a finite-horizon Markov Decision Process (MDP), which is usually described as a $5$-tuple $(\mathcal{S}, \mathcal{A}, P, R, T)$:
\begin{itemize}
    \item \(\mathcal{S}\): The state space represents the set of all possible system states. Each state \(s \in \mathcal{S}\) describes the system status of ride-hailing or ride-pooling at a specific time step.
    \item \(\mathcal{A}\): The action space $\mathcal{A}$ represents the decisions available to the match-maker at each time step: whether or not to execute a matching operation in the ride-hailing or ride-pooling system. Each action \(a \in \mathcal{A}\) directly impacts the system's state transition.
    \item \(P\): The transition function defines the transition probability $P(s'|s,a)$ from a state $s \in \mathcal{S}$ to another state $s' \in \mathcal{S}$ after taking action $a \in \mathcal{A}$. This reflects the dynamics of passengers and drivers.
    \item \(R\): The reward function \( R(s_t, a_t, s_{t+1}) \) defines the immediate reward obtained by the match-maker when executing action \( a_t \) in state \( s_t \) at time step \( t \), subsequently transitioning to the next state \( s_{t+1} \). This signal reflects the waiting time of passengers in a given state, capturing the efficiency of the ride-hailing and ride-pooling systems.
    \item \(T\): The horizon represents the total number of time steps in the finite-horizon MDP.
\end{itemize}
In a finite-horizon MDP, the match-maker interacts with the environment by observing its state $s_t \in \mathcal{S}$ and selecting an action $a_t \in \mathcal{A}$ at each time step $t = 0, \ldots, T-1$.
The objective is to maximize the expected cumulative reward (total return) 
$G$ over the finite horizon:
\begin{equation}
    G = \mathbb{E}\left[\sum_{t=0}^{T-1} R(s_t, a_t, s_{t+1})\right]
    \label{eq:G}
\end{equation}

Formulating the problem of when to match as a finite-horizon MDP captures the sequential and dynamic nature of this problem, allowing for adaptive strategies to handle real-time supply-demand fluctuations. This framework provides a foundation for applying RL algorithms to learn efficient solutions. 
The key components of MDP are given as follows. %In the following, we detail the specific design of action, state and reward function of the MDP.

\subsubsection{Action} 
The action of the match-maker is represented by a binary variable \( a_t \in \{0, 1\} \), where \( a_t = 0 \) represents skip matching, accumulating more passengers and drivers for future matches. 
\( a_t = 1 \) indicates conduct the matching, triggering the spatial matching algorithm to assign drivers to passengers.

\subsubsection{State}
Determining the optimal timing for matching in ride-hailing and ride-pooling services is a complex decision problem, which has been formulated as a finite-horizon MDP in the previous section, where the match-maker has complete knowledge of the system state at any time step. However, in practice, real-world ride-hailing and ride-pooling systems rarely provide full visibility due to challenges such as data delays, uncertainty about future events, and the complexity of system dynamics.

Given these challenges, the problem is better modeled as a Partially Observable Markov Decision Process (POMDP) \cite{kaelbling_planning_1998}, where the match-maker must make decisions based on partial observations of the system state. 
While we leave the exploration of more sophisticated state designs and the use of recurrent neural networks (RNNs) \citep{rumelhart1986learning, hochreiter_long_1997, hausknecht_deep_2017} for future work, we simplify the problem by treating it as an MDP.
This is achieved by selecting a set of relevant state variables to define the state space, enabling computationally efficient training while retaining the key elements necessary for effective decision-making.
%Theoretically, an MDP assumes that the future state depends only on the current state and the action taken, implying that the state contains all relevant information about the system dynamics. However, in practice, the information listed above may not fully satisfy this requirement. For instance:
%\begin{itemize}
   % \item Real-time geographic information and traffic data may be incomplete or delayed.
  %  \item The system lacks visibility into future passenger requests or driver availability.
   % \item The decision-making process may depend on unobserved variables, such as individual passenger behavior or market dynamics.
  %  \item It is difficult to represent complex passenger orders and drivers’ geographical distribution using low-dimensional vectors.
%\end{itemize}
%These limitations mean that the problem inherently becomes a Partially Observable Markov Decision Process (POMDP) \cite{KAELBLING199899}, where the agent does not have full visibility of the true system state.

%While a complete state representation might include all available information, utilizing such a comprehensive state is often infeasible due to practical constraints, such as computational resources, real-time data accessibility, and the high dimensionality of the input space. In this study, a simplified state representation is adopted, focusing on critical observable elements that significantly impact matching decisions. This simplification allows the framework to remain computationally efficient while retaining decision-making effectiveness.

In this study, the state at each time step \( t \) is represented as:
\begin{align}
    s_t = \left[ T_t, \Delta T_t, N_p(t), \overline{W}_p(t), W_{\max}(t), N_d(t) \right],
\end{align}
where each variable denotes a relevant statistic:
\begin{itemize}
    \item \( T_t \): Current system time, providing context for time-of-day variations.
    \item \( \Delta T_t \): Time elapsed since the last matching operation.
    \item \( N_p(t) \): Number of unmatched passenger requests in the system.
    \item \( \overline{W}_p(t) \): Average waiting time of unmatched passenger requests.
    \item \( W_{\max}(t) \): Maximum waiting time among unmatched passenger requests.
    \item \( N_d(t) \): Number of available drivers in the system.
\end{itemize}

\subsubsection{Reward}

\paragraph{Natural Reward Design}
The reward function in this RL framework aims to minimize passenger waiting time. In ride-hailing systems, total waiting time consists of matching waiting time (time until a driver is assigned) and driver arrival waiting time (time until the driver arrives). In ride-pooling, an additional detour delay due to shared routes is included.

For ride-hailing, the reward \( R^{H}(s_t, a_t, s_{t+1}) \) at state \( s_t \), after executing action \( a_t \) and transitioning to state \( s_{t+1} \), is defined as:
\begin{align}
    R^H(s_t, a_t ,s_{t+1}) = -(\phi R_m(s_t, a_t, s_{t+1}) +R_w(s_t, a_t, s_{t+1}))
    \label{eq:reward_oh}
\end{align}
where $\phi \in [0,\infty)$ is the coefficient reflecting passengers' relative tolerance for waiting to be picked up compared to the waiting to be matched. \( R_m(s_t, a_t, s_{t+1}) \) represents the incremental waiting time for all unmatched passengers at the current state \( s_t \), after taking action \( a_t \) and transitioning to state \( s_{t+1} \), which is formulated as $ R_m(s_t, a_t, s_{t+1}) = \Delta t N_p^{unmatch}(s_t, a_t)$. Here $\Delta t$ represents the length of the time step, $N_p^{unmatch}(s_t, a_t)$ represents the number of unmatched orders at state $s_t$ after taking action $a_t$. \( R_w(s_t, a_t, s_{t+1}) \) represents the waiting time for matched passengers to be picked up by drivers at the current state\( s_t \), after taking action \( a_t \) and transitioning to state \( s_{t+1} \), which is formulated as:
\begin{align}
    R_w(s_t, a_t, s_{t+1}) = 
    \begin{cases}
    0, & a_t = 0 \\
    f_{pick}(s_t), & a_t = 1
    \end{cases}
\end{align}
\label{eq:R_w}
where  \(f_{pick}(s_t)\) represents the function to calculate the sum of the waiting time for matched passengers to be picked up by drivers at the current state \( s_t \), assuming a matching decision is made. The specific calculation method of \(f_{pick}(s_t)\) will be detailed in the Matching Algorithm section. 

For ride-pooling, the reward function $R^{P}(s_t, a_t, s_{t+1})$ additionally includes detour delay \( R_d(s_t, a_t, s_{t+1}) \):
\begin{align}
    R^{P}(s_t, a_t, s_{t+1}) = -\big(&\phi R_m(s_t, a_t, s_{t+1}) + \tau R_d(s_t, a_t, s_{t+1}) \nonumber \\
    &+ R_w(s_t, a_t, s_{t+1})\big)
    \label{eq:reward_op}
\end{align}
where \(\tau\) is a another weighting coefficient reflecting passengers' higher tolerance for detour delay compared to the time spent waiting to be matched. \( R_d(s_t, a_t, s_{t+1}) \) represents the detour delay caused by ride-pooling for all matched passengers at state \(s_t\), after taking action \(a_t\) and transitioning to state \( s_{t+1} \), which is formulated as:
\begin{align}
    R_d(s_t, a_t, s_{t+1}) = 
    \begin{cases}
    0, & a_t = 0 \\
    f_{detour}(s_t), & a_t = 1
    \end{cases}
\end{align}
\label{eq:R_d}
where \(f_{detour}(s_t)\) represents the function to calculate the sum of the detour delay for matched passengers at the current state \( s_t \), assuming a matching decision is made. The specific calculation method of \(f_{detour}(s_t)\) will also be detailed in the Matching Algorithm section.

\paragraph{Reward Sparsity and Potential-based Reward Shaping}
Although the reward functions $R^H$ and $R^P$ for ride-hailing and ride-pooling are well-defined and theoretically suitable for learning an adaptive match-maker policy to determine matching decisions, they suffer from the challenge of sparse rewards.
The sparse reward problem arises because \( R_w(s_t, a_t, s_{t+1}) \) and \( R_d(s_t, a_t , s_{t+1}) \) yield non-zero feedback only when  \(  a_t = 1 \), providing no information when \(  a_t = 0 \). As a result, the match-maker receives limited guidance during training, making it difficult to learn an effective policy. 

To address this issue, we employ \textit{Potential-Based Reward Shaping} (PBRS) \cite{ng1999policy}, a method designed to refine the reward structure without altering the optimal policy. In the following, we describe how the PBRS is applied to mitigate the sparse reward problem and improve learning efficiency.  

PBRS modifies the reward function of RL with a potential function \(\Phi: \mathcal{S} \rightarrow \mathbb{R} \) intending to provide a denser learning signal for faster convergence.
The modified reward function is defined as follows in finite-horizon MDPs:
\begin{align}
    &R'(s_t, a_t, s_{t+1} ) \nonumber \\
    = &\begin{cases}
        R(s_t, a_t, s_{t{+}1}) + \Phi(s_{t{+}1}) - \Phi(s_t) & t \neq T-1, \\
        R(s_{T{-}1}, a_{T{-}1}, s_{T}) - \Phi(s_{T-1}) + \Phi(s_0), & t = T-1.
    \end{cases}
    \label{eq:pbrs_fhmdp}
\end{align}
where \( \Phi(s_t) \) reflects the desirability of state \( s_t \). 
Importantly, as shown in Equation~\eqref{eq:pbrs_fhmdp}, the modified reward at the final step $T{-}1$ differs from that of other steps. 
This adjustment is made to compensate for the discrepancy introduced by PBRS, ensuring that the expected total return $G'$ under the modified reward function \(R'\) remains identical to the expected original total return $G$ for all policies (for detailed calculations, see Equation~\eqref{eq:G}). 
As a result, PBRS preserves the optimal policy. The proof is provided below:
\begin{align}
    G' &= \mathbb{E} \left[\sum_{t=0}^{T-1} R'(s_t, a_t, s_{t{+}1}) \right] \nonumber \\
    &= \mathbb{E} \left[
        \begin{aligned}
            &\sum_{t=0}^{T-2} \Bigl(R(s_t,a_t, s_{t{+}1}) + \Phi(s_{t+1}) - \Phi(s_t)\Bigr) \\
            &+ R'(s_{T-1}, a_{T-1}, s_T)
        \end{aligned}
    \right] \nonumber \\
    &= \mathbb{E} \left[
        \begin{aligned}
            &\sum_{t=0}^{T-2} R(s_{t}, a_t, s_{t{+}1}) - \Phi(s_0) \\
            &+ \Phi(s_{T-1}) + R'(s_{T-1}, a_{T-1}, s_T)
        \end{aligned}
    \right] \nonumber \\
    &= \mathbb{E} \left[ \sum_{t=0}^{T-2} R(s_{t}, a_t, s_{t{+}1}) + R(s_{T-1}, a_{T-1}, s_T) \right] \nonumber \\
    &= \mathbb{E} \left[ \sum_{t=0}^{T-1} R(s_{t}, a_t, s_{t{+}1})  \right] \nonumber \\
    &= G \label{eq:prove_PBRS}
\end{align}
The proof above shows that incorporating a potential function into the reward structure does not alter the expected total return for any policy. Consequently, a policy that is optimal under the original reward function remains optimal under the modified reward function. Moreover, prior studies have demonstrated that, when the potential function is appropriately selected—often with the aid of domain knowledge—PBRS can significantly accelerate learning without detriment to the agent's asymptotic performance, particularly in environments characterized by sparse rewards, as in our case \citep{ng1999policy, devlin2011theoretical, ibrahim2024comprehensive, bal2022reward}. Below, we describe the design of our potential function.

Our objective is to construct a potential function that prevents the rewards from being concentrated solely on the matching step, as is the case with the original rewards. Instead, the modified rewards should be distributed more uniformly across all actions in the sequence. This uniform distribution enables the match-maker to better discern advantageous states, thereby accelerating the learning process. In addressing the problem of determining when to match in ride-hailing and ride-pooling systems, the potential function must accurately reflect the desirability of the current state with respect to the optimization objective. Given that our ultimate goal is to minimize passenger waiting time, which is closely related to the matching outcome, we can hypothesize a matching action at the current time step and evaluate the result of this hypothetical match. By constructing a potential function based on the quality of this hypothetical matching result, we can better guide the match-maker in optimizing its strategy. The quality of matching results in ride-hailing and ride-pooling can be directly evaluated using the previously mentioned \(f_{pick}(s_t)\) and \(f_{detour}(s_t)\): the former calculates the time passengers wait for a matched driver to arrive, while the latter measures the detour delay for passengers in ride-pooling.

Below are our designs of potential functions in the cases of ride-hailing and ride-pooling:
\begin{align}
    \Phi^h(s_t) &= -f_{pick}(s_t) \label{eq:potential_h} \\
    \Phi^p(s_t) &= -(f_{pick}(s_t)+f_{detour}(s_t)) \label{eq:potential_p}
\end{align}
The functions \(f_{pick}(s_t)\) and \(f_{detour}(s_t)\) are originally used to calculate \(R_w\) and \(R_d\), respectively, and are only evaluated when \(a_t = 1\). However, by introducing them into the potential function, they can now be evaluated for all actions, regardless of whether matching occurs (\(a_t = 1\)) or not (\(a_t = 0\)). This design allows the potential function to provide a consistent evaluation of the current state, offering insights into the hypothetical impact of performing a matching operation.

% Combining the definition of original reward functions in equation \eqref{eq:reward_oh} and equation \eqref{eq:reward_op} and the potential functions in equation \eqref{eq:potential_h} and \eqref{eq:potential_p}, we have our final reward functions after applying PBRS as follows:
% \begin{align}
%     R'^{H}(s_t, a_t, s_{t{+}1}) = 
%     \begin{cases}
%         R^{H}(s_t, a_t, s_{t{+}1}) +  \Phi^h(s_{t+1}) - \Phi^h(s_t), \hfill t \neq T-1, \\
%         R^{H}(s_{T-1}, a_{T-1}, s_{T}) \\
%         \quad - \Phi^h(s_{T-1}) + \Phi^h(s_0), \hfill t = T-1.
%     \end{cases}
% \end{align}

% \begin{align}
%     R'^{P}(s_t, a_t, s_{t{+}1}) = 
%     \begin{cases}
%         R^{P}(s_t, a_t, s_{t{+}1}) + \Phi^p(s_{t+1}) \\
%         \quad - \Phi^p(s_t), \hfill t \neq T-1, \\
%         R^{P}(s_{T-1}, a_{T-1}, s_{T}) \\
%         \quad - \Phi^p(s_{T-1}) + \Phi^p(s_0), \hfill t = T-1.
%     \end{cases}
% \end{align}

\begin{figure}[h]
    \centering
    \includegraphics[width=\linewidth]{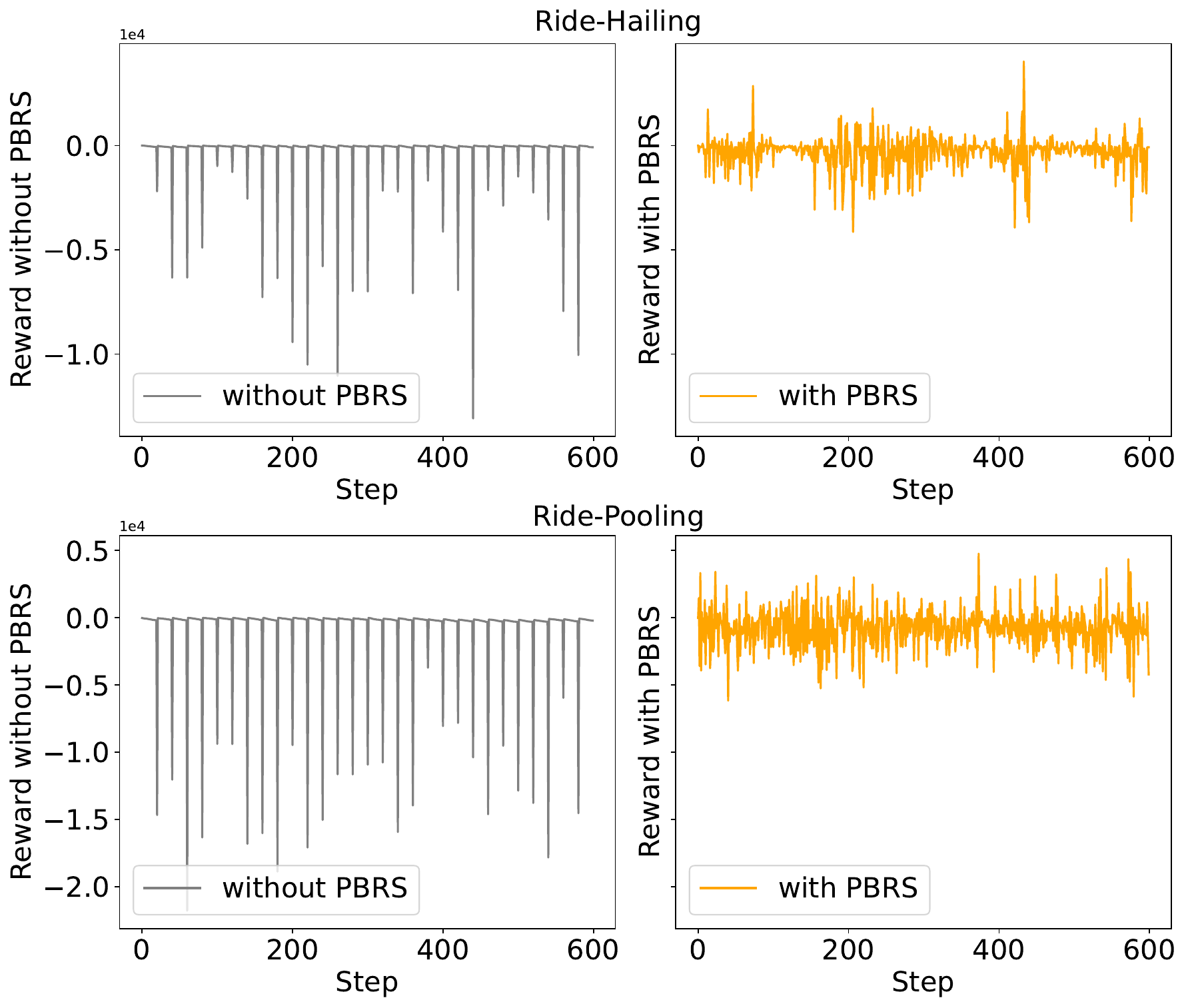} 
    \caption{Comparison of Rewards without and with PBRS} 
    \label{fig:reward_compare} 
\end{figure}

Figure~\ref{fig:reward_compare} examines the distribution of rewards without (Equations \ref{eq:reward_oh}{,}~\ref{eq:reward_op} and with PBRS (Equations \ref{eq:pbrs_fhmdp}{,}~\ref{eq:potential_h}{,}~\ref{eq:potential_p}). The gray line (without-PBRS) exhibits sparse rewards with large negative fluctuations, which can hinder training convergence. In contrast, the orange line (with-PBRS) shows smoother and more frequent rewards, effectively mitigating the sparse reward issue and providing consistent feedback.

This design maintains the overall learning signal while offering the agent more frequent feedback, encouraging better decisions during waiting and matching. Consequently, PBRS enhances the training efficiency and performance of the ride-hailing and ride-pooling systems.

\subsection{Learning Algorithm}

This study employs the Proximal Policy Optimization (PPO) algorithm \cite{schulman2017ppo} to determine the matching timing in ride-hailing and ride-pooling services. PPO is widely recognized for its stability, computational efficiency, and robust performance in dynamic environments like ride-hailing. Its clipped objective function ensures stable learning by preventing overly large policy updates, making it particularly suitable for environments with fluctuating supply and demand. Additionally, PPO's scalability and adaptability enable it to perform effectively across a wide range of scenarios, optimizing decision-making in large-scale simulations.

PPO improves upon traditional policy gradient methods such as REINFORCE \cite{williams1992simple} and A2C \cite{mnih2016asynchronous} by introducing a surrogate objective function that constrains updates to the policy. Specifically, PPO uses the following clipped objective to optimize the policy:
\begin{align}
    L^{\text{CLIP}}(\theta) = \mathbb{E}_t \left[ \min \left( r_t(\theta) \hat{A}_t, \text{clip}(r_t(\theta), 1-\epsilon, 1+\epsilon) \hat{A}_t \right) \right]
\end{align}

Where \( r_t(\theta) = \frac{\pi_\theta(a_t|s_t)}{\pi_{\theta_{\text{old}}}(a_t|s_t)} \) is the probability ratio between the new policy \( \pi_\theta \) and the old policy \( \pi_{\theta_{\text{old}}} \); \( \hat{A}_t \) is the estimated advantage function, which measures the relative value of action \( a_t \) in state \( s_t \); \( \epsilon \) is a hyperparameter (typically 0.1 or 0.2) that controls the range of permissible updates to the policy.

The clipping mechanism ensures that the probability ratio \( r_t(\theta) \) remains within a safe range \([1-\epsilon, 1+\epsilon]\), preventing large updates that could destabilize learning.

In addition to the policy update, PPO uses a learned critic \cite{sutton1999policy} to estimate state values and advantages. The total loss function combines the policy loss \( L^{\text{CLIP}} \), the value loss for the critic, and an entropy term to encourage exploration:
\begin{align}
    L(\theta) = L^{\text{CLIP}}(\theta) - c_1 L^{\text{VALUE}}(\theta) + c_2 L^{\text{ENTROPY}}(\theta)
\end{align}

Where \( L^{\text{VALUE}}(\theta) = \left( V_\theta(s_t) - V_t^{\text{target}} \right)^2 \) minimizes the difference between the estimated and target state values; \( L^{\text{ENTROPY}}(\theta) \) encourages policy entropy, promoting diverse action selection; \( c_1 \) and \( c_2 \) are coefficients controlling the relative importance of value loss and entropy.

In this study, PPO trains an agent (match-maker) to decide when to match based on passenger demand, driver availability, and system delays. The neural network architecture is based on an actor-critic framework with multi-layer perceptrons (MLPs). Both the actor and critic networks consist of three hidden layers with 64 units per layer, activated by the Tanh function. The actor network outputs action probabilities, while the critic estimates state values. Actions are sampled from a Bernoulli distribution.

The agent’s performance is tracked to monitor key metrics, such as passenger delays. By determining the matching timing, this framework demonstrates significant improvements in system efficiency, effectively addressing the challenges of dynamic ride-hailing and ride-pooling environments.

The basic flow of the PPO algorithm implemented in this study is shown in the pseudocode below. And details of the training parameters are provided in the Appendix.

\begin{algorithm}
\caption{PPO Algorithm}
\begin{algorithmic}[1]
\For{episode = 1 to $E$}
    \State Reset the environment and get initial state $s_0$
    \State Initialize episode return $Return = 0$ and $done = False$
    \For{$t = 0$ to $T-1$}
        \State Sample action $a_t$ from policy $\pi_\theta(a_t|s_t)$
        \State Execute action $a_t$ in environment
        \State Observe reward $r_t$, next state $s_{t+1}$, and done status $d_t$
        \State Store $(s_t, a_t, r_t, s_{t+1}, d_t)$ in trajectory buffer
        \If{$done$}
            \State \textbf{break}
        \EndIf
    \EndFor

    \State Compute discounted returns $R_t$ and advantages $\hat{A}_t$ using GAE
    \For{$k = 1$ to $K$} 
        \State Sample minibatch from trajectory buffer
        \State Compute importance ratio $r_t(\theta) = \frac{\pi_\theta(a_t|s_t)}{\pi_{\theta_{\text{old}}}(a_t|s_t)}$
        \State Compute clipped surrogate objective:
        \State $L^{\text{CLIP}}(\theta) = \mathbb{E}_t \left[ \min \left( r_t(\theta) \hat{A}_t, \text{clip}(r_t(\theta), 1\pm\epsilon) \hat{A}_t \right) \right]$
        \State Compute value function loss: 
        \State $L^{\text{VALUE}}(\theta) = \left( V_\theta(s_t) - V_t^{\text{target}} \right)^2$
        \State Compute entropy bonus: 
        \State$L^{\text{ENTROPY}}(\theta) = \mathbb{E}_t \left[ -\pi_\theta(a_t|s_t) \log \pi_\theta(a_t|s_t) \right]$
        \State Compute total loss: 
        \State$L(\theta) = L^{\text{CLIP}}(\theta) - c_1 L^{\text{VALUE}}(\theta) + c_2 L^{\text{ENTROPY}}(\theta)$
        \State Update policy and value function using gradient ascent on $L(\theta)$
    \EndFor
\EndFor
\end{algorithmic}
\end{algorithm}

This pseudocode summarizes the PPO algorithm’s structure, highlighting policy optimization, value estimation, and gradient updates, which allow the agent to effectively learn strategies for optimizing matching intervals.

\section{Simulator Design}
To validate and evaluate the proposed approach, we develop a simulator to model the dynamics of ride-hailing and ride-pooling services. The simulator utilizes a real public dataset of taxi trips in Manhattan, New York City \cite{nyc_tlc_trip_data}, to generate realistic passenger orders and driver distributions, capturing both spatial and temporal fluctuations in supply and demand. Details on the generation passenger and driver data process are provided in Subsection A. 
Also, the simulator incorporates matching algorithms for both ride-hailing and ride-pooling settings, as described in Subsection B. 

To ensure a balance between realism and computational efficiency, several assumptions are made. Vehicles are assumed to travel at a constant speed of 40 km/h to simplify travel time calculations. Passengers cancel their orders if unmatched within five minutes, and drivers wait at pick-up points for up to 10 minutes before repositioning. Additionally, pooling is limited to two passenger orders per vehicle.

%When the match-maker decides to perform matching at a specific time step, vehicles are assigned to passengers based on factors such as geographic proximity and waiting time. This setup enables the match-maker to learn the optimal timing for matching to minimize passenger waiting time. %Moreover, the simulation environment serves as a robust testing platform, facilitating the evaluation of optimal matching strategies to enhance the efficiency of ride-hailing and ride-pooling services.
\subsection{Passenger and Driver Data Generation}
\label{data generation}
% The simulator leverages historical for-hire vehicle (FHV) trip records provided by the New York City government \cite{nyc_tlc_trip_data}, which include request times, pickup and drop-off details, and origin-destination zones. A Poisson distribution is utilized to model the stochastic nature of ride requests and driver availability, capturing temporal and spatial variations in demand and supply. Each geographical zone is assigned a unique Poisson distribution with a calibrated mean rate ($\lambda$) derived from historical data, enabling realistic event generation that varies by time and location.

The simulator leverages historical for-hire vehicle (FHV) trip records provided by the New York City government \cite{nyc_tlc_trip_data}, specifically from January to March 2022. These records include request times, pickup and drop-off details, and predefined origin-destination zones. The zones are classified according to the structure already provided in the dataset, reflecting the geographic divisions used in the original records. To model the stochastic nature of ride requests and driver availability, a Poisson distribution is employed, capturing temporal and spatial variations in demand and supply. Each zone is assigned a unique Poisson distribution with a calibrated mean rate ($\lambda$) calculated as the average number of requests per time period, derived from the historical data. This approach enables realistic event generation that reflects time- and location-specific demand patterns.

% Passenger destinations are represented using a transition probability matrix constructed from historical trip data. This matrix encodes the likelihood of travel between zones, incorporating observed patterns such as commuting routes and peak demand areas, thereby ensuring realistic traffic flow simulation.

Passenger destinations are represented using a transition probability matrix constructed from historical trip data. This matrix encodes the likelihood of travel between geographical zones, capturing observed patterns such as commuting routes and peak demand areas to ensure realistic traffic flow simulation. Each entry in the matrix represents the proportion of trips originating from a specific zone and ending in another zone. This is calculated as the ratio of trips between the two zones to the total number of trips originating from the same zone, providing a detailed and data-driven representation of passenger movement within the city.

\begin{figure}[h]
    \centering
    \includegraphics[width=\linewidth]{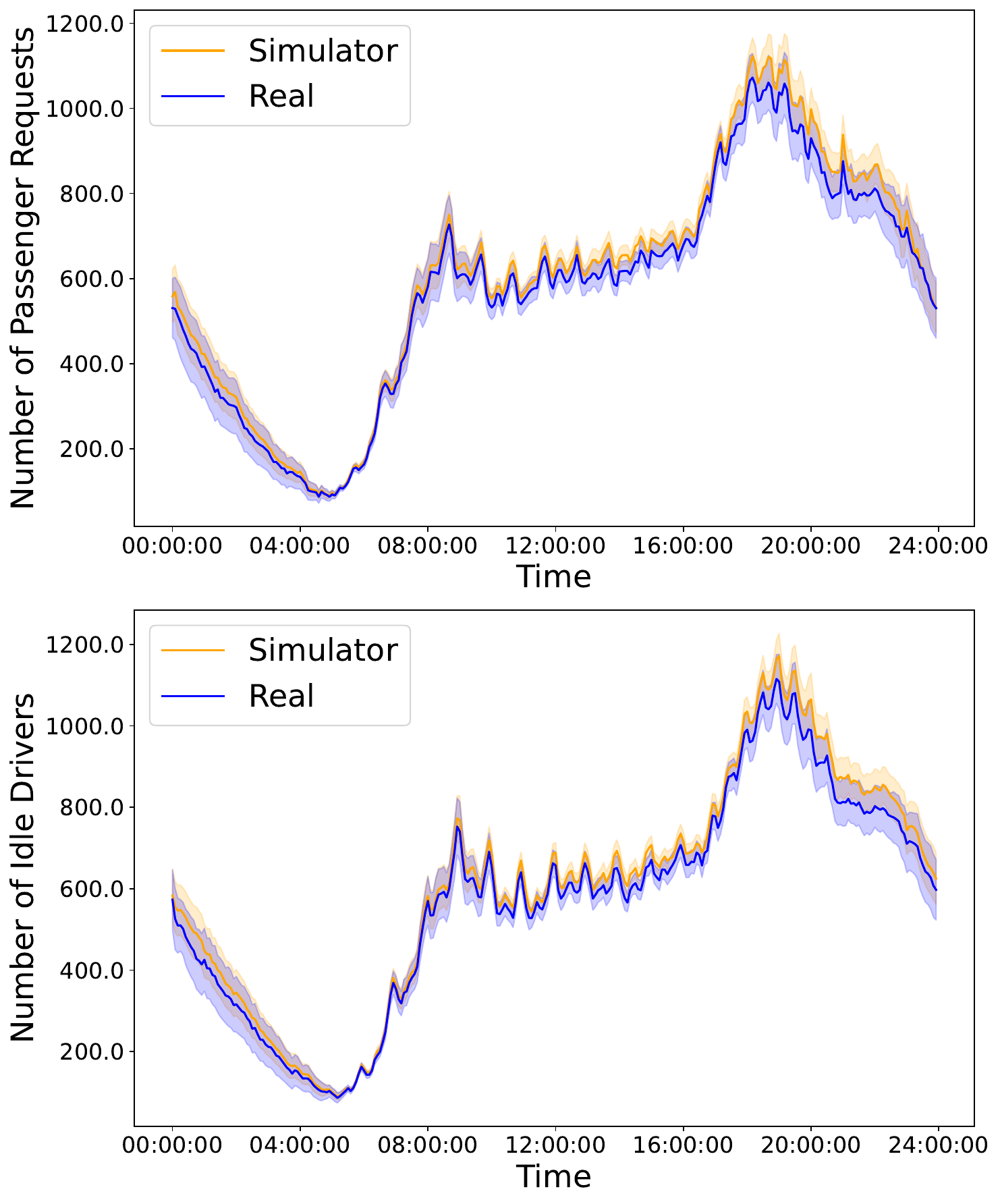} 
    \caption{Comparison of Passenger Request Data and Idle Driver Data Generated by the Simulator and Real Data} 
    \label{fig:sampled_data_compare} 
\end{figure}

The generated dataset comprises passenger request and cancellation times, pickup points, destinations, driver locations, and availability, creating a dynamic, second-by-second simulation environment. As depicted in Figure~\ref{fig:sampled_data_compare}, the simulated passenger request data and idle driver data closely matches real-world daily variations in ride demand and supply. This setup supports the reinforcement learning framework by providing varied conditions for training and evaluation, with continuous decision points driven by fluctuating demand and supply. Parameter adjustments allow for diverse scenario testing, ensuring the algorithm's robustness and adaptability to real-world conditions.

\subsection{Matching Algorithms}
This subsection describes the matching algorithms designed for ride-hailing and ride-pooling services. For ride-pooling, the matching process consists of two stages: first, orders with similar origins and destinations are paired through passenger-passenger matching; second, the paired passengers are matched with a driver. For ride-hailing, the passenger-driver matching process is similar to ride-pooling, except it involves a single passenger being matched with a driver instead of a passenger pair. 

\subsubsection{Passenger-Passenger Matching (ride-pooling)}
In the simulation, the passenger-passenger matching algorithm matches passenger orders with the goal of minimizing the detour distance. To quantify the impact of detour on travel efficiency, we define detour delay rate, which is formulated as: 
\[
\text{DDR} = \frac{\text{Distance without detour}}{\text{Distance after detour}}
\]
where "Distance without detour" represents the direct travel distance for a single passenger, and "Distance after detour" is the actual distance when accommodating multiple passengers.  Higher DDR values correspond to more efficient pairings with minimal detour effects, while lower values indicate significant disruptions.

\begin{figure}[h]
    \centering   \includegraphics[width=\linewidth]{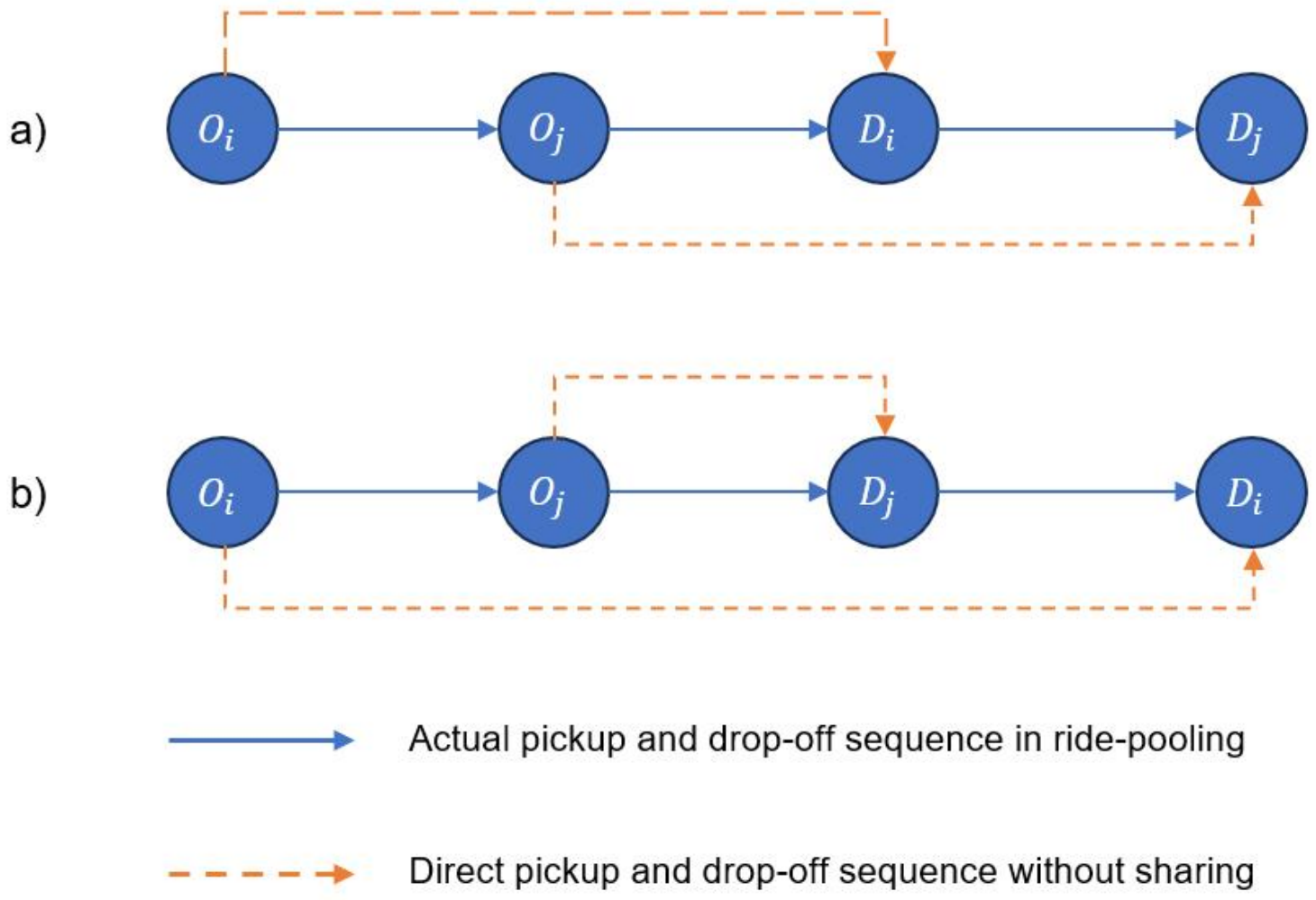} 
   \caption{Possible Pickup and Drop-off Sequences for Two Passenger Orders} 
    \label{fig:DDR_route} 
\end{figure}

However, when calculating the DDR values for two matched passenger orders \(i\) and \(j\), their DDR values are usually different. Additionally, different pickup and drop-off sequences may result in varying DDR values. As shown in Figure~\ref{fig:DDR_route}, the pickup and drop-off sequences can be categorized into two cases, labeled as (a) and (b) in the figure. Here, \(O_i\) and \(D_i\) represent the origin and destination of passenger \(i\), respectively. 

In case (a), both orders \(i\) and \(j\) experience detours due to the ride-pooling route. For order \(i\), the distance after the detour is increased due to the detour through \(O_j\). Similarly, for order \(j\), the distance after the detour is increased due to the detour through \(O_i\). Therefore, the DDR values for orders \(i\) and \(j\) under pickup and drop-off sequence (a), denoted as \(DDR^a(i)\) and \(DDR^a(j)\), can be computed as follows:
\begin{align}
    DDR^a(i) = \frac{d(O_i, D_i)}{d(O_i, O_j, D_i)}
\end{align}
\begin{align}
    DDR^a(j) = \frac{d(O_j, D_j)}{d(O_j, D_i, D_j)}
\end{align}
Where \(d(\cdot)\)  represents the shortest path distance in a road network when traveling through a specified sequence of locations. To evaluate the matching quality between orders \(i\) and \(j\) in case (a), we define \(DDR^a(i,j)\) as the smaller value between \(DDR^a(i)\) and \(DDR^a(j)\). This represents the DDR of the passenger in the pair who is more significantly affected by the detour. The corresponding formula is as follows:

\begin{align}
    DDR^a(i,j) = \min (\frac{d(O_j, D_j)}{d(O_j, D_i, D_j)},\frac{d(O_j, D_j)}{d(O_j, D_i, D_j)})
\end{align}

In case (b), the situation is more straightforward. As shown in Figure~\ref{fig:DDR_route}, only order \(i\) experiences a detour, while order \(j\) is directly delivered from its origin to its destination. Consequently, the DDR value for order \(j\), \(DDR^b(j)\), is equal to 1. To evaluate the matching quality in case (b), \(DDR^b(i,j)\) is determined by the DDR value of order \(i\), \(DDR^b(i)\). The corresponding formula is as follows:

\begin{align}
    DDR^b(i,j) = \frac{d(O_i, D_i)}{d(O_i, O_j, D_j, D_i)}
\end{align}

Therefore, by calculating the DDR values for the two pickup and drop-off sequences, the optimal sequence can be selected, which corresponds to the sequence with the larger DDR value. The DDR value of the optimal sequence is then used to assess the matching quality between orders \(i\) and \(j\), denoted as \(DDR(i,j)\). The formula for computing \(DDR(i,j)\) is as follows:

    \begin{equation}
        \text{DDR}(i, j) = \max \left( 
        \begin{array}{l}
            \min \left( 
                \frac{d(O_i, D_i)}{d(O_i, O_j, D_i)}, \frac{d(O_j, D_j)}{d(O_j, D_i, D_j)} 
            \right), \\[5pt]
            \frac{d(O_i, D_i)}{d(O_i, O_j, D_j, D_i)}
        \end{array}
        \right)
        \label{eq:DDR} 
    \end{equation}

Once the pickup and drop-off sequence between orders \(i\) and \(j\) has been determined, the detour distance for order \(i\), \(\Delta d_i\),  caused by the ride-pooling can also be calculated. The corresponding formulas are as follows:

\begin{align}
    \Delta d_i = d(O_i, \dots, D_i \mid \sigma^*) - d(O_i, D_i)
    \label{eq:detour} 
\end{align}

where  \(d(O_i, D_i)\) is the shortest path distance for passenger \(i\) if served independently; \(d(O_i, \dots, D_i \mid \sigma^*)\) is the actual travel distance from \(O_i\) to \(D_i\) under the optimal pickup and drop-off sequence \(\sigma^*\).

%The DDR guides the algorithm to prioritize matches with higher values, balancing operational efficiency with passenger satisfaction by keeping detour effects manageable.
%\begin{figure}[h]
    %\centering
    %\includegraphics[width=\linewidth]{fig/four_rout.pdf} 
   % \caption{All Possible Pickup and Drop-off Sequences for Two Passenger Orders} 
    %\label{fig:four_route} 
%\end{figure}

%The passenger-passenger matching process involves structured steps to evaluate and optimize potential matches, focusing on maximizing successful matches and minimizing detour impact.

So, the passenger-passenger matching process comprises three steps:
\begin{enumerate}
    \renewcommand{\labelenumi}{\alph{enumi}.}
    \item Retrieve Passenger Orders: At each time step, the algorithm retrieves all active passenger orders, including information such as origin, destination, and request time. 
    \item Evaluate Matches: By iterating over all active passenger orders, all feasible passenger-passenger pairs are enumerated. For each feasible pair, the matching quality and the resulting detour for each order after matching are calculated based on equations \eqref{eq:DDR} and \eqref{eq:detour}.
    \item Optimize Passenger-passenger Pairing: To handle cases where orders can be matched with multiple others, we optimize pairing as an integer linear optimization problem, represented by a graph \( G = (P_{s_t}, E_{s_t}) \), where \( P_{s_t} \) is the set of passenger orders at state \( s_t \), and \( E_{s_t} \) represents the set of feasible passenger pairs. Each edge \( e \in E_{s_t} \) has a weight \( \omega(e) \), which reflects the matching utility (based on the Detour Delay Rate, DDR). The decision variable \( x(e) \) is binary, where \( x(e) = 1 \) indicates that the pair corresponding to edge \( e \) is selected. 

    The optimization problem is formulated as:
    \begin{align}
        & \max \sum_{e \in E_{s_t}} \omega(e) \cdot x(e), \\
        & \sum_{e \in \delta(p)} x(e) \leq 1, & \forall p \in P_{s_t}, \\
        & x(e) \in \{0, 1\}, & \forall e \in E_{s_t}.
    \end{align}
    
    where, \( \delta(p) \) denotes the set of edges connected to passenger order \( p \). The first constraint ensures that each passenger can only be matched with one other passenger. Solving this optimization problem yields the set of optimal passenger-passenger pairs \( PP_{s_t} \) at state \( s_t \), which maximizes the total DDR.

\end{enumerate}

%\begin{figure}[h]
    %\centering
    %\includegraphics[width=\linewidth]{fig/visualize_route_pp_match.pdf} 
  %  \caption{Passenger-Passenger Matching Result Illustration} 
   % \label{fig:PP_match_result} 
%\end{figure}

%Figure~\ref{fig:PP_match_result} visualizes the results of passenger-passenger matching, demonstrating that this spatial matching algorithm effectively pairs passengers with similar routes and plans optimal pick-up and drop-off paths, minimizing detours for passengers.

Based on the optimal matching results, the sum of the detour delay for matched passengers at the current state \( s_t \) can be calculated:

\begin{align}
    f_{detour}(s_t) = \sum_{i \in PP_{s_t}} \Delta d_i / v
\end{align}

Where \(v\) represents the average vehicle speed.

The passenger-passenger matches are then treated as single entities in the Passenger-Driver Match phase, allowing for seamless integration into the overall ride-hailing system.

\subsubsection{Passenger-Driver Matching}
This phase finalizes the matching process for regular ride-hailing services and serves as the second stage for ride-pooling, where passengers (or passenger pairs in ride-pooling) are assigned to the nearest available driver to minimize pickup waiting time. The driver-passenger matching problem is modeled as an Integer Linear Programming (ILP) problem, formulated as:
\begin{align}
    & \min \sum_{i \in D_{s_t}} \sum_{j \in PP_{s_t}} T_{ij} \cdot x_{ij}, \\
    & \sum_{i \in D_{s_t}} x_{ij} \leq 1, \\
    & \sum_{j \in PP_{s_t}} x_{ij} \leq 1, \\
    & x_{ij} \in \{0, 1\}, & \forall i \in D_{s_t}, \ \forall j \in PP_{s_t},
\end{align}
where:
$D_{s_t}$ is the set of available drivers at state \(s_t\). \(PP_{s_t}\) is the set of unmatched passenger orders (ride-hailing) or passenger pairs (ride-pooling) at \(s_t\).
\(T_{ij}\) is the pickup time between driver \(i \in D_{s_t}\) and passenger (or passenger pair) \(j \in PP_{s_t}\), estimated as $T_{ij} = \frac{d_{ij}}{v}$, where \(d_{ij}\) is the distance between driver \(i\) and passenger (or passenger pair) \(j\). \(x_{ij}\): A binary decision variable, where \(x_{ij} = 1\) if driver \(i\) is assigned to passenger (or passenger pair) \(j\), and \(x_{ij} = 0\) otherwise. Constraints (37) and (38) indicate that each passenger (or passenger pair) can be matched to at most one driver. And each driver can serve at most one passenger (or passenger pair). The optimal result of this optimization problem, $x^*_{ij}$, is used to calculate \(f_{\text{pick}}(s_t)\), which represents the total pickup waiting time at state \(s_t\). Mathematically, we have:
\[
f_{\text{pick}}(s_t) = \min \sum_{i \in D_{s_t}} \sum_{j \in PP_{s_t}} T_{ij} \cdot x^*_{ij},
\]

%The first constraint ensures that each passenger or passenger pair can be matched to at most one driver, while the second constraint ensures that each driver can serve at most one passenger or passenger pair. Solving this optimization problem yields the optimal driver-passenger assignments for state \(s_t\), minimizing total pickup time.

%\begin{figure}[h]
   % \centering
    %\includegraphics[width=\linewidth]{fig/visualize_route_RideHailing.pdf} 
  %  \caption{Passenger-Driver Matching Result Illustration} 
    \label{fig:PD_match_result} 
%\end{figure}

%Figure~\ref{fig:PP_match_result} visualizes the results of passenger-driver matching, showing that this algorithm efficiently matches passengers to nearby available drivers and plans optimal routes for drivers, thereby reducing passenger waiting time.

%The result of the optimization problem is equivalent to \(f_{\text{pick}}(s_t)\), which represents the total pickup waiting time at state \(s_t\). Mathematically, we have:
%\[
%f_{\text{pick}}(s_t) = \min \sum_{i \in D_{s_t}} \sum_{j \in PP_{s_t}} T_{ij} \cdot x_{ij},
%\]

\section{Experiments}
In this section, extensive experiments and sensitivity analyses
are conducted to evaluate the performance of the proposed
methods
\subsection{Experimental Settings, Baselines and Evaluation Metrics} 

The match-maker was trained and evaluated using the simulator we designed. Each \textit{episode} represents 10 minutes of service, consisting of 600 discrete time steps (one second per time step). During each episode, the match-maker interacted with the simulator to decide when to perform matching operations. Its performance was assessed across multiple episodes to ensure robustness and generalizability under diverse conditions. 

\begin{figure*}[h]
    \centering
    \includegraphics[width=0.9\linewidth]{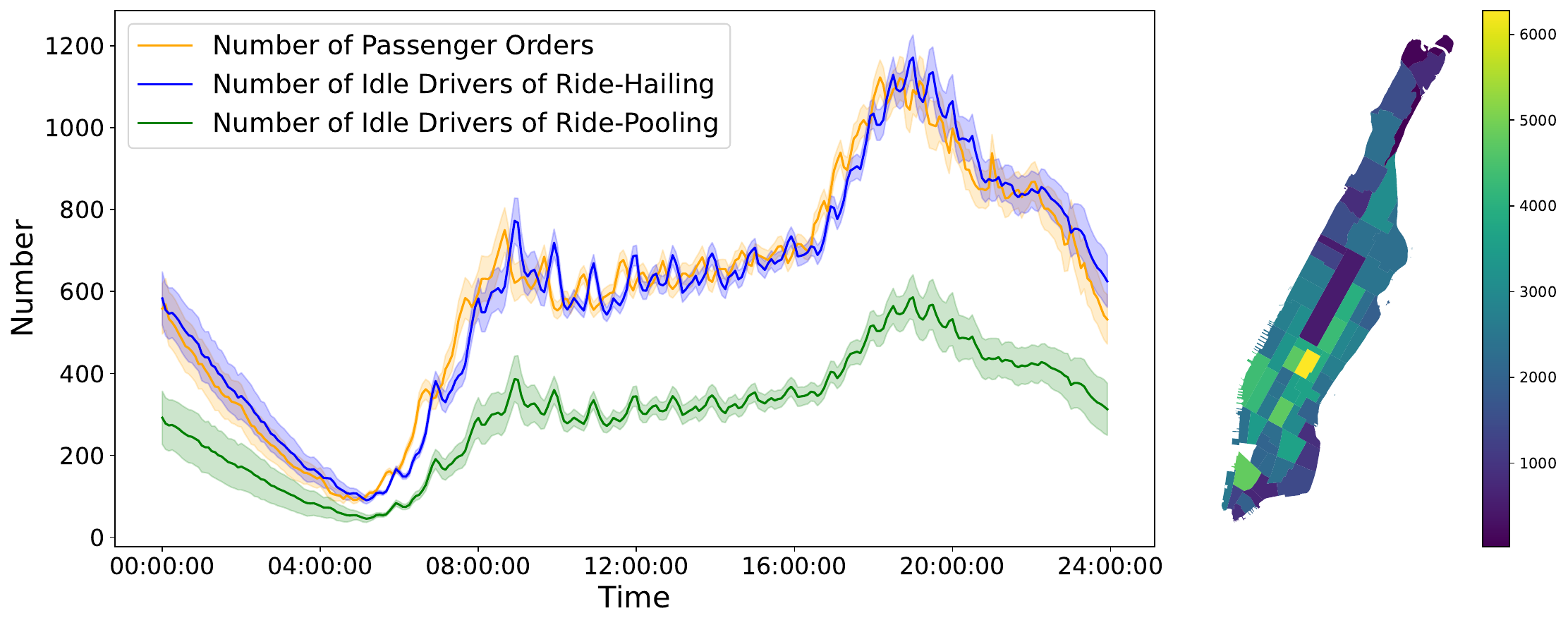} 
    \caption{Daily Supply-Demand Fluctuations and Geographical Distribution in Ride-Hailing and Ride-Pooling Systems} 
    \label{fig:/Trend_in_Passenger_Order} 
\end{figure*}

Passenger orders and driver data for each episode were generated using the method described in the section \ref{data generation}. This method is based on historical FHV trip records from Manhattan, New York City, spanning from January to March 2023 \cite{nyc_tlc_trip_data}. The dataset reflects the fluctuations in supply and demand, as well as the geographical distribution, throughout the day for both ride-hailing and ride-pooling systems, as shown in Figure~\ref{fig:/Trend_in_Passenger_Order}. 

To evaluate the effectiveness of our approach, we compare it against the \textbf{batched matching} strategy with fixed time intervals, a strategy currently implemented by ride-hailing platforms such as Uber, serves as the baseline for our study. In the ride-hailing simulator, the optimal batched matching interval is fixed at 15 seconds, while in the ride-pooling simulator, it is set to 20 seconds. These intervals were selected based on testing within the same simulator to ensure they provide a representative evaluation across various performance metrics. In the subsequent part of experiments, the \textbf{first dispatch} strategy will also be compared with the proposed approach. It is important to note that, since the simulation runs in discrete time steps, the first dispatch strategy implies that matching occurs at every time step within the simulator.

To evaluate the performance of our approach, we designed the following evaluation metrics: 
\begin{itemize}
    \item \textbf{Average Pickup Time}: The average pick-up waiting time per passenger after being matched.
    \item \textbf{Average Matching Time}: The average waiting time from request to successful matching per passenger.
    \item \textbf{Average Detour Delay}(only for ride-pooling): The average delay time per passenger due to detours.
    \item \textbf{Average Total Waiting Time}: The overall average waiting time per passenger.
    \item \textbf{Total Pickup Time}: The total time spent waiting for drivers to pick up passengers.
    \item \textbf{Total Matching Time}: The total time spent waiting for matching to occur across all passengers.
    \item \textbf{Total Detour Delay}(only for ride-pooling): The total delay caused by detours in ride-pooling.
    \item \textbf{Total Waiting Time}: The cumulative waiting time of all passengers, measuring overall waiting.
\end{itemize}

%The main body of this study focuses on the average metrics, while the cumulative metrics are detailed in the appendix for reference and analysis.

%The experiments comprehensively evaluated the RL strategy. Experiment 1 examined training performance and the role of Potential-Based Reward Shaping (PBRS) in enhancing learning efficiency. Experiment 2 compared the RL strategy with other strategies, focusing on system efficiency. Experiment 3 is designed to further investigate the operational mechanisms of the RL strategy developed in this study. Experiment 4 conducted a cross-comparison of ride-hailing and ride-pooling under different scenarios, exploring the difference between two travel modes. These experiments demonstrated the effectiveness of the RL strategy in improving system efficiency and user experience in dynamic urban environments.

The experiments are designed to answer the following questions:
\begin{itemize}
    \item \textbf{Section \ref{Training Performance}} How does the proposed RL approach converge during training for ride-hailing and ride-pooling services, and what is the impact of Potential-Based Reward Shaping on learning efficiency and performance? And how does the proposed RL strategy evolve during training?

    \item \textbf{Section \ref{Comparison of Strategies}} What are its performance advantages compared to first dispatch and batched matching strategies (including baseline) in ride-hailing and ride-pooling services? 
    
    \item \textbf{Section \ref{Detail of RL Strategy}} What is the \textit{adaptability} of the proposed approach in response to real-time supply-demand variations in ride-hailing and ride-pooling systems?
    
   % \item \textbf{Section \ref{Ride-hailing vs Ride-pooling}}  What are the differences between ride-hailing and ride-pooling under the same supply-demand conditions? What changes occur after applying the RL strategy?
    
    % \item \textbf{Section \ref{Ride-hailing vs Ride-pooling}}  \textcolor{red}{What are the key differences of applying the designed approach to ride-hailing and ride-pooling? }
\end{itemize}

\subsection{Training Performance}
\label{Training Performance}

This experiment evaluates the training performance of the RL strategy for Ride-Hailing and Ride-Pooling services during peak traffic hours (8:30–8:40 a.m.) in Manhattan. Each service mode was trained over more than 4,800 episodes (2,880,000 steps), reflecting a highly dynamic supply-demand environment. Passenger requests and driver locations were generated using a Poisson distribution, with parameters fitted and estimated based on the historical data mentioned before. Variability across episodes was introduced by probabilistically sampling different passenger and driver datasets. It is important to note that the datasets of passengers and drivers used in each episode for the experiments in this and subsequent sections are independently generated and unique. The same episodes are not reused across experiments.

To evaluate the impact of Potential-Based Reward Shaping (PBRS), two reward designs—one with PBRS and one without—were compared. Each training experiment was repeated five times with different random seeds to ensure reliability, generalizability, and reproducibility. Results were averaged, and confidence intervals were calculated to account for randomness and ensure statistical reliability. The baseline strategy was tested over 1,000 random episodes under identical simulation settings for fair comparisons. 
\begin{figure}[h]
    \centering
    \includegraphics[width=\linewidth]{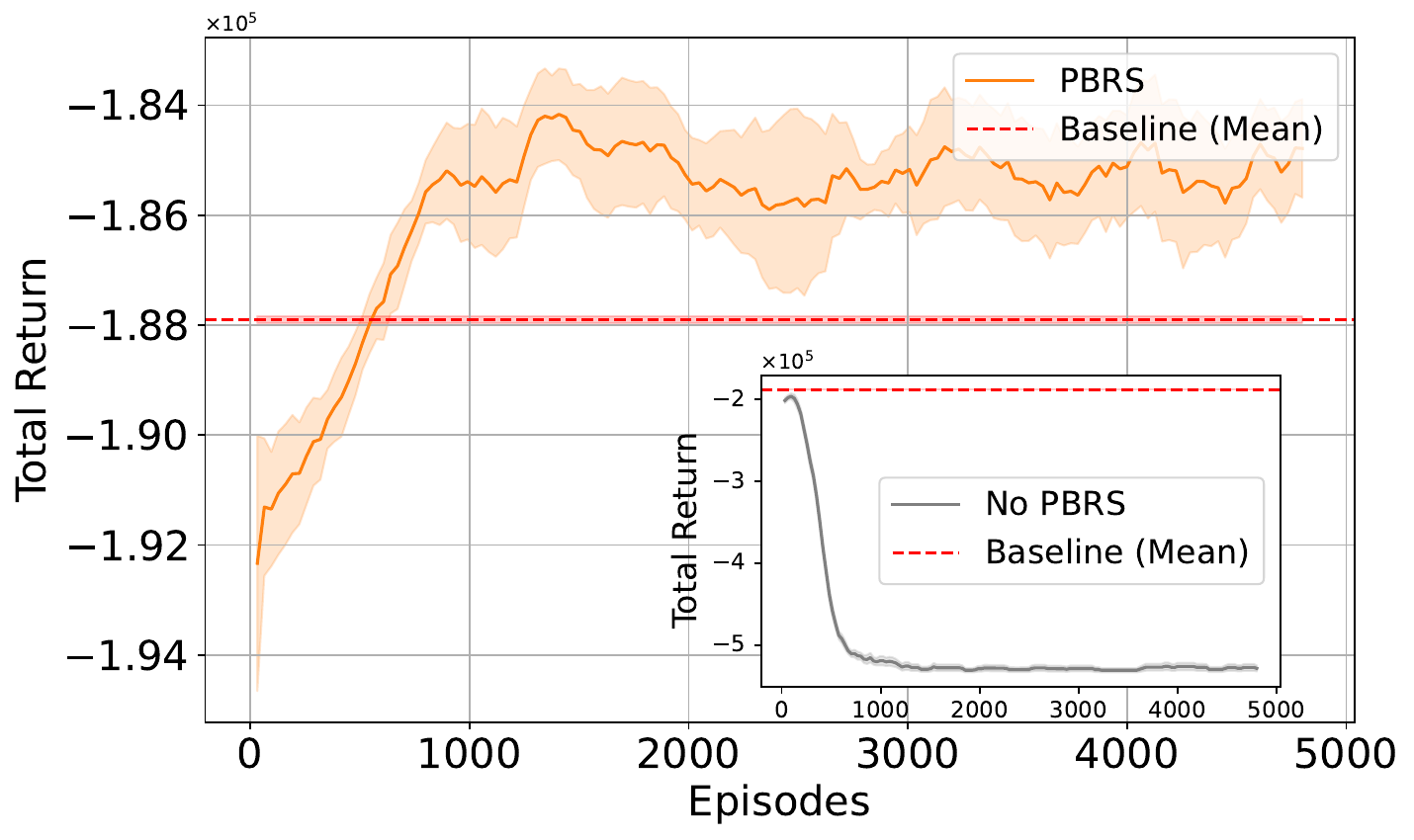} 
    \caption{Training Curve of Traditional Ride-Hailing Service} 
    \label{fig:/traing_curve_ride_hailing} 
\end{figure}

Figure~\ref{fig:/traing_curve_ride_hailing} illustrates the Total Return progression during training for the traditional ride-hailing service with and without PBRS. The gray curve (without PBRS) shows that, in the initial few hundred episodes, the Total Return does not improve with training. Instead, it decreases rapidly before stabilizing at a low level in the later stages. Upon analyzing the actions taken by the agent, it was observed that the issue of sparse rewards prevented the agent from consistently receiving 
\( R_w(s_t, a_t) \) signals. As a result, in the later stages of training, the agent repeatedly chose to wait instead of initiating matching actions. This behavior caused passenger waiting times to increase continuously, further lowering the overall system performance. In contrast, the orange curve, representing the PBRS-enhanced strategy, demonstrates faster convergence and significantly higher Total Return, stabilizing around -186,000 after approximately 600 episodes. PBRS mitigates the sparse reward issue by providing more frequent feedback, enabling the agent to make better decisions and improve learning efficiency. The confidence intervals for the PBRS curve narrow as training progresses, reflecting reduced variability and improved stability in the strategy. 
\begin{figure}[h]
    \centering
    \includegraphics[width=\linewidth]{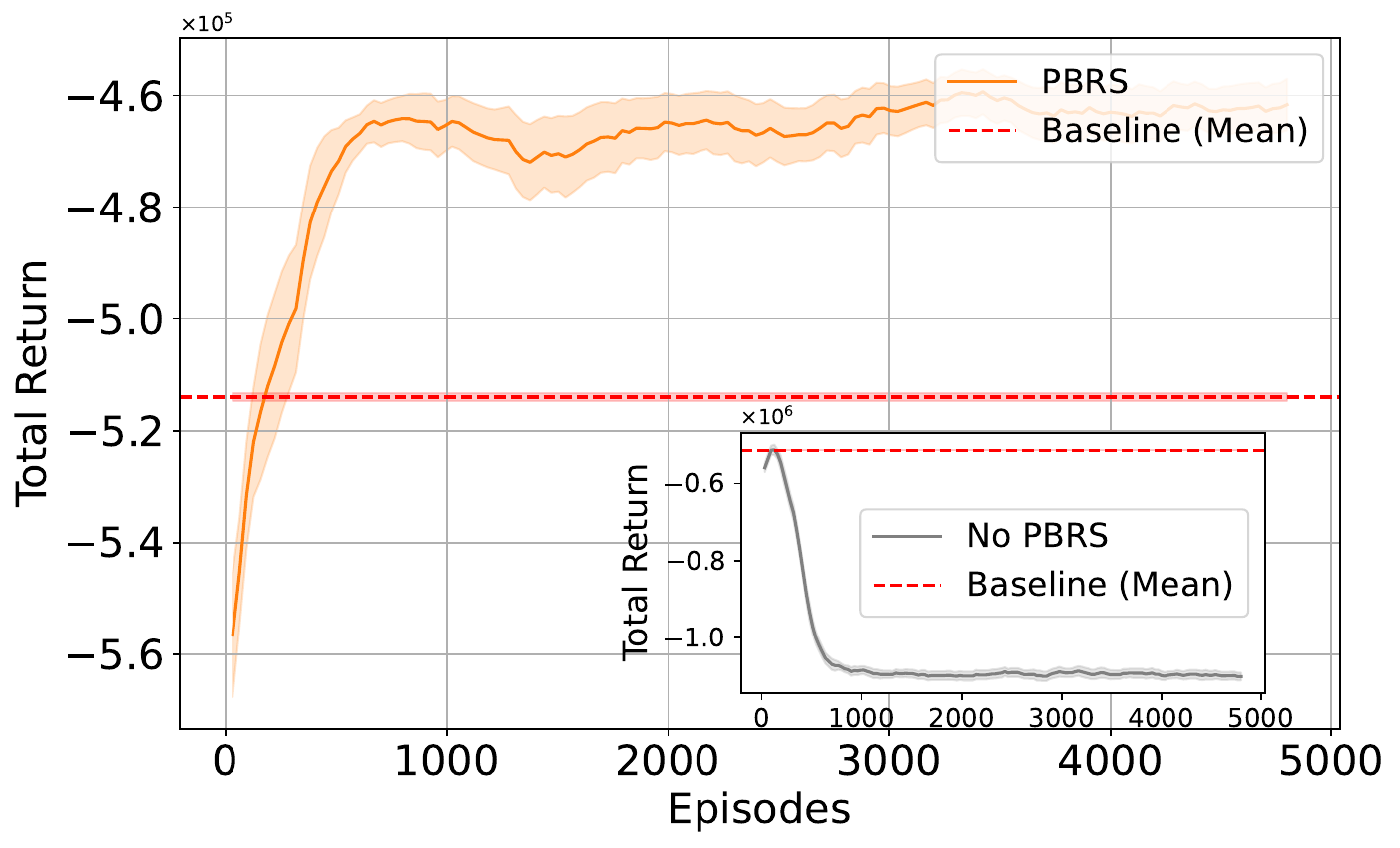} 
    \caption{Training Curve of Ride-Pooling Service} 
    \label{fig:/traing_curve_Shared} 
\end{figure}

The same trend can also be found in the Ride-Pooling service, as shown in Figure~\ref{fig:/traing_curve_Shared}. The gray curve, representing the non-PBRS strategy, exhibits a sharp decline in Total Return during the initial episodes, followed by stabilization at a very low level, far below the baseline of -520,000. In contrast, the orange curve, representing the PBRS-enhanced strategy, shows a rapid improvement in Total Return during the first 500 episodes. It stabilizes at approximately -460,000 after 1,000 episodes, well above the baseline level. The frequent feedback provided by PBRS helps the agent learn more efficiently, improving its ability to handle complex multi-passenger scenarios. Early in training, the PBRS curve displays wider confidence intervals, reflecting variability due to exploration. However, as training progresses, the confidence intervals narrow, indicating that the strategy becomes more stable and consistent.

Figure~\ref{fig:/traing_curve_ride_hailing} and Figure~\ref{fig:/traing_curve_Shared} show that PBRS improves the learning process in both ride-hailing and ride-pooling environments, achieving Total Return levels consistently above the baseline in both cases. The ride-pooling curve stabilizes with narrower confidence intervals, reflecting more consistent performance in the learned strategy. In contrast, traditional ride-hailing shows more fluctuations after convergence, which could be due to the simpler matching conditions inherent in this service mode. Additionally, the rapid initial increase in ride-pooling Total Return highlights its effectiveness in optimizing multi-passenger matching during the early stages of training.

\begin{figure}[h]
    \centering
    \includegraphics[width=\linewidth]{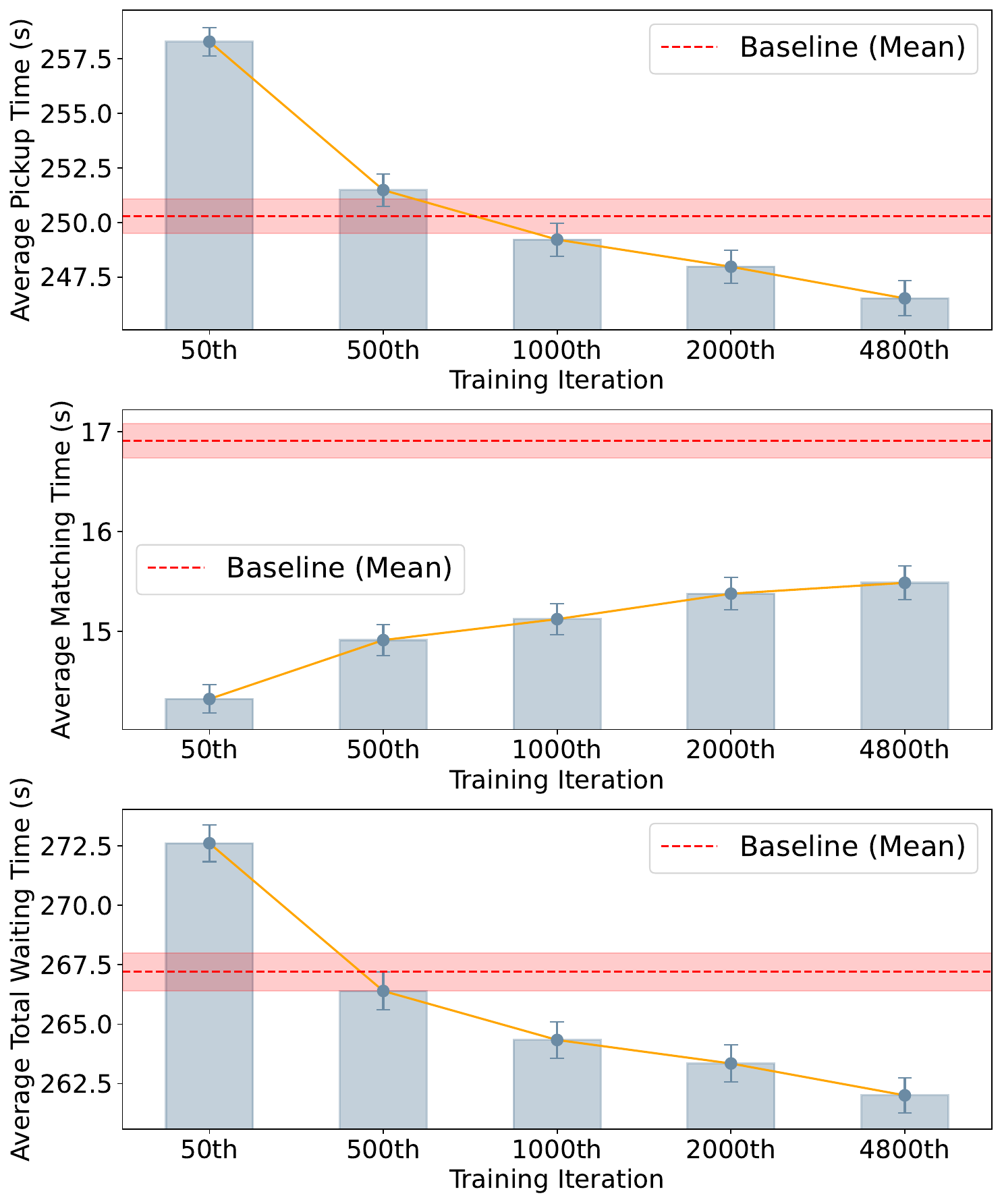} 
    \caption{Comparison of Metrics Across Training Episodes and Baseline Strategy for Ride-Hailing Services} 
    \label{fig:/metrics_ride_hailing} 
\end{figure}

\begin{figure}[H]
    \centering
    \includegraphics[width=0.8\linewidth]{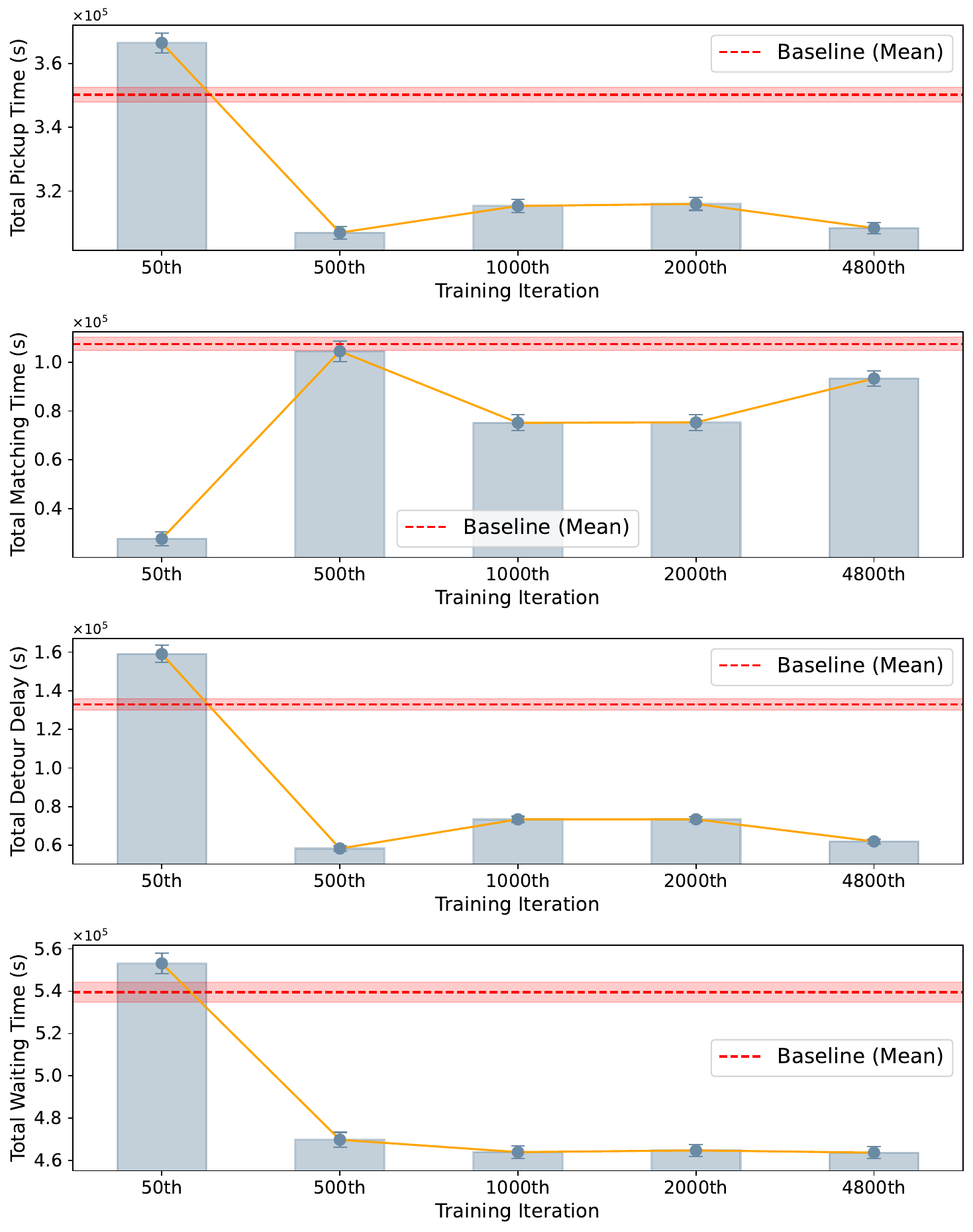} 
    \caption{Comparison of System Waiting Time Across Training Episodes and Baseline Strategy for Ride-Pooling Services} 
    \label{fig:/metrics_shared_ride_hailing_system} 
\end{figure}

To further analyze the evolution of the strategy during training, we selected five training checkpoints on the PBRS-enhanced curve (at episode 50, 500, 1000, 2000, and 4800), representing different learning stages from initial exploration to convergence. The strategies at these checkpoints, along with the baseline strategy, were tested over 1,000 episodes in the same simulation environment to ensure fair comparison. Key performance metrics were recorded to evaluate the changes and improvements in the RL strategy across different training stages. The results are shown in Figures~\ref{fig:/metrics_ride_hailing} and~\ref{fig:/metrics_shared_ride_hailing}, with system evaluation metrics detailed in Appendix Figures~\ref{fig:/metrics_ride_hailing_system} and~\ref{fig:/metrics_shared_ride_hailing_system}.

Figure~\ref{fig:/metrics_ride_hailing} demonstrates that as training progresses, the Average Pickup Time decreases from 258 seconds at the early stages to 247 seconds by the 4800th episode. Although the reduction in Pickup Time per passenger is modest, it results in significant system-wide improvements, highlighting the effectiveness of the trained strategy in optimizing Pickup Time. In contrast, the baseline strategy shows limited capacity to reduce Pickup Time.

The Average Matching Time increases slightly from 14.4 seconds to 15.7 seconds during training. This suggests that the trained strategy prioritizes pickup optimization, leading to minor trade-offs in Matching Time. Despite this increase, the Matching Time remains lower than the baseline strategy.

The Average Total Waiting Time per Passenger decreases from 270 seconds to 262 seconds, which is lower than the baseline strategy's 267 seconds. This indicates that the trained strategy effectively balances the trade-off between Pickup Time and Matching Time, ultimately reducing passenger waiting times across the entire ride-hailing matching process.

\begin{figure}[h]
    \centering
    \includegraphics[width=\linewidth]{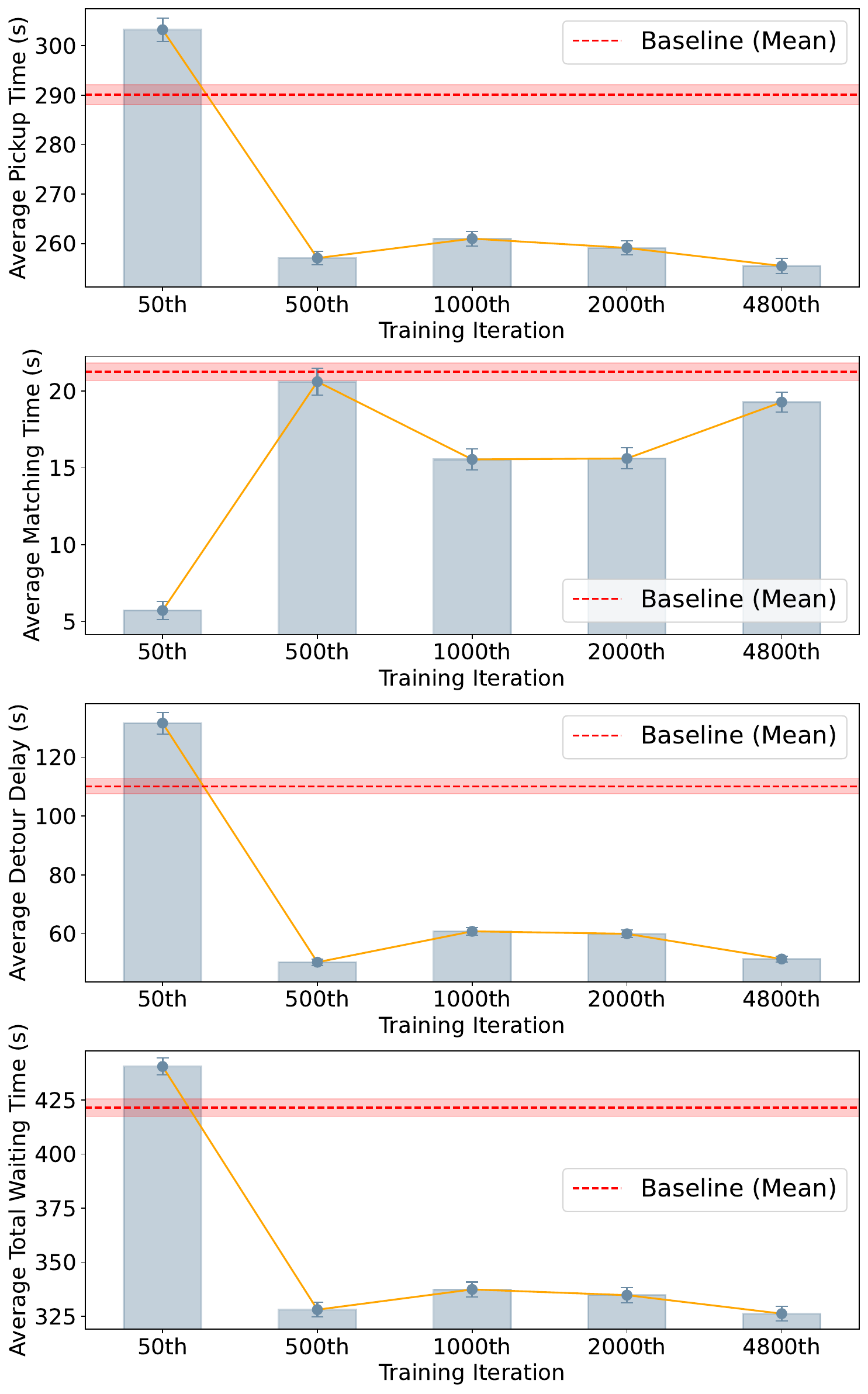} 
    \caption{Comparison of Metrics Across Training Episodes and Baseline Strategy for Ride-Pooling Services} 
    \label{fig:/metrics_shared_ride_hailing} 
\end{figure}

In ride-pooling, the trained strategy must balance Pickup Time, Detour Delay, and Matching Time to further reduce Total Waiting Time. The results in Figure~\ref{fig:/metrics_shared_ride_hailing} provide a clear visualization of this process.

The Average Pickup Time decreases from approximately 310 seconds at episode 50 to 250 seconds by episode 4800, stabilizing around this value, while the baseline remains higher at 290 seconds, reflecting improved efficiency in reducing post-matching waiting time. The Detour-related metric shows a similar substantial improvement. Average Detour Delay decreases from 125 seconds to 55 seconds, compared to the baseline's 110 seconds.

The Average Matching Time increases during training, peaking at 20 seconds before stabilizing around 18 seconds by episode 4800, outperforming the baseline's 22 seconds. This trend highlights the trained strategy’s ability to handle complex ride-pooling scenarios, albeit with slightly longer matching times due to the increased complexity of decision-making, compared to ride-hailing.

With Pickup Time, Detour Delay, and Matching Time all lower than the baseline, the Average Total Waiting Time naturally shows improvement. The Average Total Waiting Time decreases from 440 seconds to 330 seconds by episode 4800, while the baseline remains at 420 seconds.

In summary, the PBRS-enhanced RL strategy, for both ride-hailing and ride-pooling, achieves better matching results by slightly extending the Matching Time, thereby effectively reducing the Total Waiting Time. Compared to the baseline strategy with fixed matching intervals, the PBRS-enhanced RL strategy not only delivers superior matching result but also achieves shorter Matching Time, resulting in a lower Total Waiting Time overall.

\subsection{Performance comparison between different matching strategies}
\label{Comparison of Strategies}

%The RL strategy trained with PBRS was evaluated against fixed matching interval strategies to assess its performance. A total of 250 random episodes were tested using identical episode settings as in the previous experiment, and various evaluation metrics were recorded (
We compare the performance of the proposed approach with batched matching and first dispatch strategies. For batched matching, we adjust the matching interval to examine its impact on performance, using 5, 15, 30, and 60 seconds for ride-hailing, and 10, 20, 40, and 80 seconds for ride-pooling, as shown in Tables~\ref{tab:strategy_comparison_ride_hailing} and~\ref{tab:strategy_comparison_ride_pooling}, respectively.

\begin{table*}[]
\centering
\caption{Performance Comparison of Different Matching Strategies in Ride-hailing}
\label{tab:strategy_comparison_ride_hailing}
\begin{tabularx}{\linewidth}{lrrrrr}
\hline
Strategies                & Average Pickup Time (s) & Average Matching Time (s) & Average Detour Delay (s) & Average Total Waiting Time (s) \\ \hline
First Dispatch  & 300.661\(\pm\)1.851     & 1.704\(\pm\)0.366         & -                        & 302.365\(\pm\)1.619           \\
5s (Fixed Time Interval)  & 276.801\(\pm\)1.821     & 6.807\(\pm\)0.331         & -                        & 283.608\(\pm\)1.564           \\
15s (Fixed Time Interval) & 251.992\(\pm\)1.601     & 16.691\(\pm\)0.317        & -                        & 268.683\(\pm\)1.322           \\
30s (Fixed Time Interval) & 245.571\(\pm\)1.574     & 31.122\(\pm\)0.308        & -                        & 276.693\(\pm\)1.211           \\
60s (Fixed Time Interval) & 235.422\(\pm\)1.411     & 62.533\(\pm\)0.310        & -                        & 297.955\(\pm\)1.198           \\
Our approach                       & 247.176\(\pm\)1.629     & 15.306\(\pm\)0.318        & -                        & \textbf{262.482\(\pm\)1.452}          \\ \hline
\end{tabularx}
\end{table*}

\begin{table*}[]
\centering
\caption{Performance Comparison of Different Matching Strategies in Ride-pooling}
\label{tab:strategy_comparison_ride_pooling}
\begin{tabularx}{\linewidth}{lrrrrr}
\hline
Strategies                & Average Pickup Time (s) & Average Matching Time (s) & Average Detour Delay (s) & Average Total Waiting Time (s) \\ \hline
First Dispatch & 416.293$\pm$1.367      & 2.011$\pm$0.568          & 190.514$\pm$2.011        & 608.818$\pm$2.099              \\
10s (Fixed Time Interval) & 355.919$\pm$1.311      & 12.054$\pm$0.455          & 145.355$\pm$1.919        & 513.328$\pm$1.876              \\
20s (Fixed Time Interval) & 291.899$\pm$1.285      & 22.038$\pm$0.373          & 113.098$\pm$1.755        & 427.035$\pm$1.670              \\
40s (Fixed Time Interval) & 280.719$\pm$1.233      & 47.178$\pm$0.371          & 102.827$\pm$1.571        & 430.724$\pm$1.322              \\
80s (Fixed Time Interval) & 265.113$\pm$1.301      & 92.519$\pm$0.398          & 75.422$\pm$1.249         & 433.054$\pm$1.317              \\
Our approach                        & 257.586$\pm$0.934      & 19.512$\pm$0.433          & 60.712$\pm$0.918         & \textbf{337.810$\pm$1.249}               \\ \hline
\end{tabularx}
\end{table*}

The comparative results presented in Table II highlight the effectiveness of the proposed approach in minimizing Average Total Waiting Time, achieving the lowest value of 262.482 seconds among all strategies evaluated. This performance demonstrates an improvement over both the first dispatch strategy and batched matching configurations with fixed intervals. While the average pickup time (247.176 seconds) for the proposed approach is slightly higher than that of the 30-second and 60-second batched strategies, this is Compensated by a significant reduction in average matching time, which is reduced to 15.306 seconds. These findings indicate that the proposed approach achieves a better trade-off between pickup time and matching time, ultimately leading to superior performance in reducing total waiting time.

Similarly, Table~\ref{tab:strategy_comparison_ride_pooling} confirms the advantages of the proposed approach in ride-pooling. While the 20-second fixed interval strategy performs best among the batched matching strategies, the proposed approach achieves a lower average total waiting time of 337.810 seconds. Additionally, it outperforms all other strategies in Average Pickup Time, Average Matching Time, and Average Detour Delay, further demonstrating its effectiveness in optimizing system performance across multiple metrics.

\begin{figure*}[h]
    \centering
    \includegraphics[width=\linewidth]{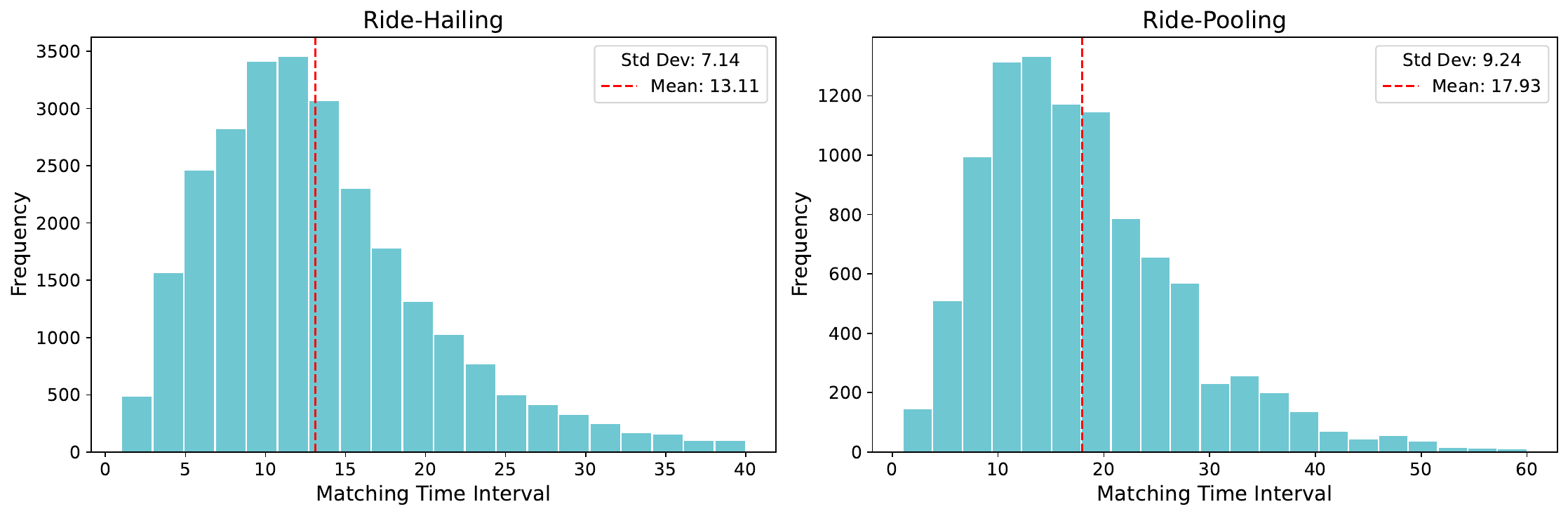} 
    \caption{Distribution of Dynamic Time Intervals} 
    \label{fig:/Distribution_interval} 
\end{figure*}

\begin{figure*}[h]
    \centering
    \includegraphics[width=\linewidth]{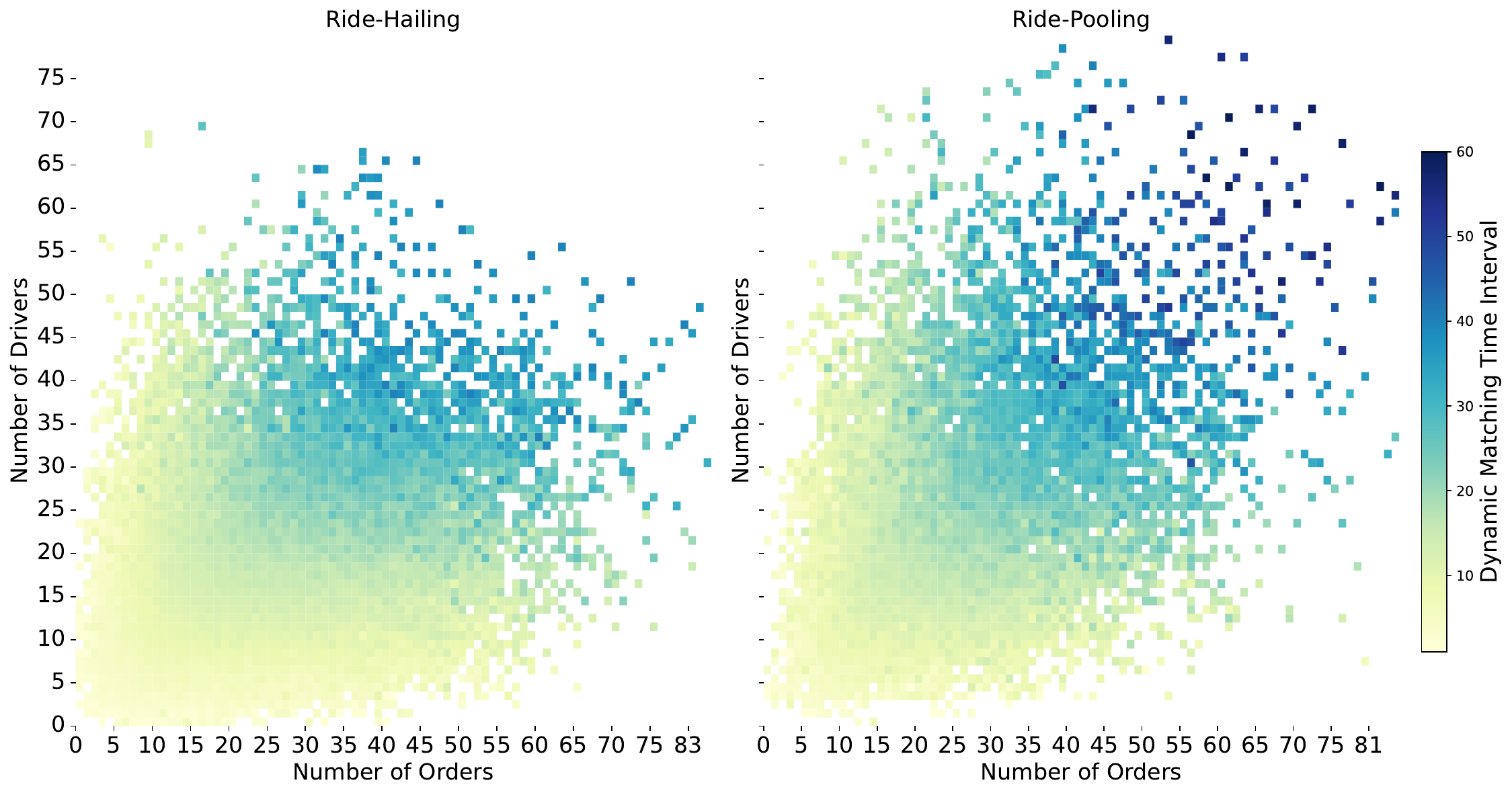} 
    \caption{Matching Interval Dynamics with Order and Driver Counts} 
    \label{fig:/Matching_Interval_Dynamics_with_Order_Driver_Counts} 
\end{figure*}

\subsection{Adaptability}
\label{Detail of RL Strategy}

The experiment is designed to test the adaptability of the RL strategy by analyzing how it adjusts matching intervals dynamically in response to fluctuating supply-demand conditions for ride-haling and ride-pooling services.

Figure~\ref{fig:/Distribution_interval} shows the distribution of dynamic matching intervals for ride-hailing and ride-pooling systems. In ride-hailing, intervals are mostly concentrated between 10–20 seconds but occasionally extend to 40 seconds. For ride-pooling, intervals are more dispersed and often exceed 30 seconds to accumulate sufficient requests for effective pooling, highlighting the complexity of coordinating multiple passengers.

Figure~\ref{fig:/Matching_Interval_Dynamics_with_Order_Driver_Counts} provides further insights into how matching intervals respond to supply-demand conditions. In ride-hailing (left chart), intervals extend as passenger orders increase, particularly when orders exceed 30. Conversely, intervals shorten when driver supply is limited, prioritizing reduced waiting times. This indicates a preference for longer intervals in balanced supply-demand conditions and shorter ones under tight supply.
In ride-pooling (right chart), intervals are generally longer, especially when passenger orders exceed 50, as the system delays matching to maximize pooling efficiency. However, under imbalanced conditions, shorter intervals are used to address supply-demand gaps promptly. 

These observations demonstrate that the proposed approach effectively adjusts matching intervals based on real-time supply and demand conditions. It extends intervals to optimize matching decisions when the system is balanced and uses shorter intervals to handle imbalances and minimize waiting times.

\section{Conclusion and Future Work}

The objective of this study is to leverage reinforcement learning techniques, specifically a dynamic strategy based on the Proximal Policy Optimization (PPO) algorithm, to determine the optimal matching timing in ride-hailing and ride-pooling services. The study introduces a dynamic optimization framework that continuously guides the match-maker on when to perform matching to reduce passenger waiting time. To address sparse rewards, Potential-Based Reward Shaping is employed, accelerating convergence to optimal strategies. This study also addresses the multi-passenger matching complexity in ride-pooling by implementing an algorithm that reduces detour delay, improving passenger satisfaction.

Through comprehensive simulation experiments, this study systematically validates the effectiveness of the PPO-based dynamic matching strategy in ride-hailing systems. The experimental results reveal three significant advantages: (1) a substantial reduction in passenger waiting time, (2) accelerated learning efficiency through PBRS implementation, and (3) enhanced pick-up and detour management capabilities in both ride-hailing and ride-pooling scenarios. Notably, the proposed strategy demonstrates remarkable robustness under varying supply-demand conditions, consistently outperforming conventional fixed-interval matching approaches. These findings not only confirm the superior flexibility and operational efficiency of the PPO-based method compared to traditional solutions but also provide valuable insights for promoting sustainable development in intelligent transportation systems.

The findings suggest that the developed reinforcement learning framework can significantly enhance urban mobility services by dynamically perform matching at the optimal timing in response to real-time fluctuations in supply and demand, thereby reducing waiting times and detours while improving user satisfaction. This study presents a practical application for intelligent optimization in current platforms like Uber and Lyft, proposing a pathway to improved service efficiency and environmental sustainability.

Future work could incorporate real-time traffic data, simulate hybrid environments, and extend ride-pooling to multiple passengers, better reflecting real-world conditions. Exploring alternative reinforcement learning algorithms and enhancing state representation through graph-based structures or advanced feature engineering may further refine decision-making and scalability, advancing intelligent optimization in urban mobility.

\appendices
\section{Parameters of PPO}
The Proximal Policy Optimization (PPO) algorithm in this project is controlled by a variety of parameters, which allow for flexible configuration and tuning to optimize performance. Below is a description of each parameter and its role in the PPO implementation.

\begin{itemize}
    \item \textbf{exp\_name}: The name of this experiment, which defaults to the name of the script.
    \item \textbf{seed}: Sets the random seed for the experiment, ensuring reproducibility. \textbf{(Default: 10)}
    \item \textbf{torch\_deterministic}: If enabled, sets \texttt{torch.backends.cudnn.deterministic = False} for deterministic operations in PyTorch. \textbf{(Default: True)}
    \item \textbf{cuda}: If enabled, CUDA (GPU) will be used by default to accelerate computation. \textbf{(Default: True)}
    \item \textbf{track}: Enables tracking of the experiment using Weights and Biases. \textbf{(Default: True)}
    \item \textbf{wandb\_project\_name}: Sets the project name in Weights and Biases. \textbf{(Default: "cleanRL")}
    \item \textbf{wandb\_entity}: Specifies the entity (team) name for the Weights and Biases project. \textbf{(Default: "baoyiman")}
    \item \textbf{capture\_video}: If enabled, videos of the agent’s performance are saved to the \texttt{videos} folder. \textbf{(Default: False)}

    \item \textbf{env\_id}: Specifies the environment ID, used to load the appropriate environment (\texttt{Ride\_hailing}).
    \item \textbf{learning\_rate}: The learning rate for the optimizer, set at \textbf{2.5e-4}.
    \item \textbf{num\_envs}: Defines the number of parallel game environments. \textbf{(Default: 4)}
    \item \textbf{num\_steps}: The number of steps taken in each environment per policy rollout. \textbf{(Default: 120)}
    \item \textbf{anneal\_lr}: Enables learning rate annealing for both policy and value networks. \textbf{(Default: True)}
    \item \textbf{gamma}: The discount factor $\gamma$ for future rewards. \textbf{(Default: 1)}
    \item \textbf{gae\_lambda}: Lambda for Generalized Advantage Estimation (GAE). \textbf{(Default: 0.95)}
    \item \textbf{num\_minibatches}: Defines the number of mini-batches per update. \textbf{(Default: 8)}
    \item \textbf{update\_epochs}: Specifies the number of epochs (K) to update the policy. \textbf{(Default: 4)}
    \item \textbf{norm\_adv}: Toggles advantage normalization for the PPO update. \textbf{(Default: True)}
    \item \textbf{clip\_coef}: Sets the surrogate clipping coefficient to prevent large policy updates. \textbf{(Default: 0.2)}
    \item \textbf{clip\_vloss}: Enables a clipped loss function for the value function, as specified in the PPO paper. \textbf{(Default: True)}
    \item \textbf{ent\_coef}: The coefficient for the entropy term to encourage exploration. \textbf{(Default: 0.01)}
    \item \textbf{vf\_coef}: The coefficient for the value function term. \textbf{(Default: 0.5)}
    \item \textbf{max\_grad\_norm}: Sets the maximum gradient norm for gradient clipping. \textbf{(Default: 1)}
    \item \textbf{target\_kl}: The target threshold for the Kullback-Leibler (KL) divergence. If exceeded, the policy update halts. \textbf{(Default: None)}

    \item \textbf{batch\_size}: The batch size, computed at runtime.
    \item \textbf{minibatch\_size}: The mini-batch size, computed at runtime.
    \item \textbf{num\_iterations}: The number of training iterations, calculated at runtime.
\end{itemize}

These parameters enable control over the PPO algorithm’s behavior, including exploration, stability, and computational efficiency, thereby facilitating effective training of the agent in the specified environment.

\section{Comparison of System Strategy Performance Metrics}
\begin{figure}[H]
    \centering
    \includegraphics[width=0.8\linewidth]{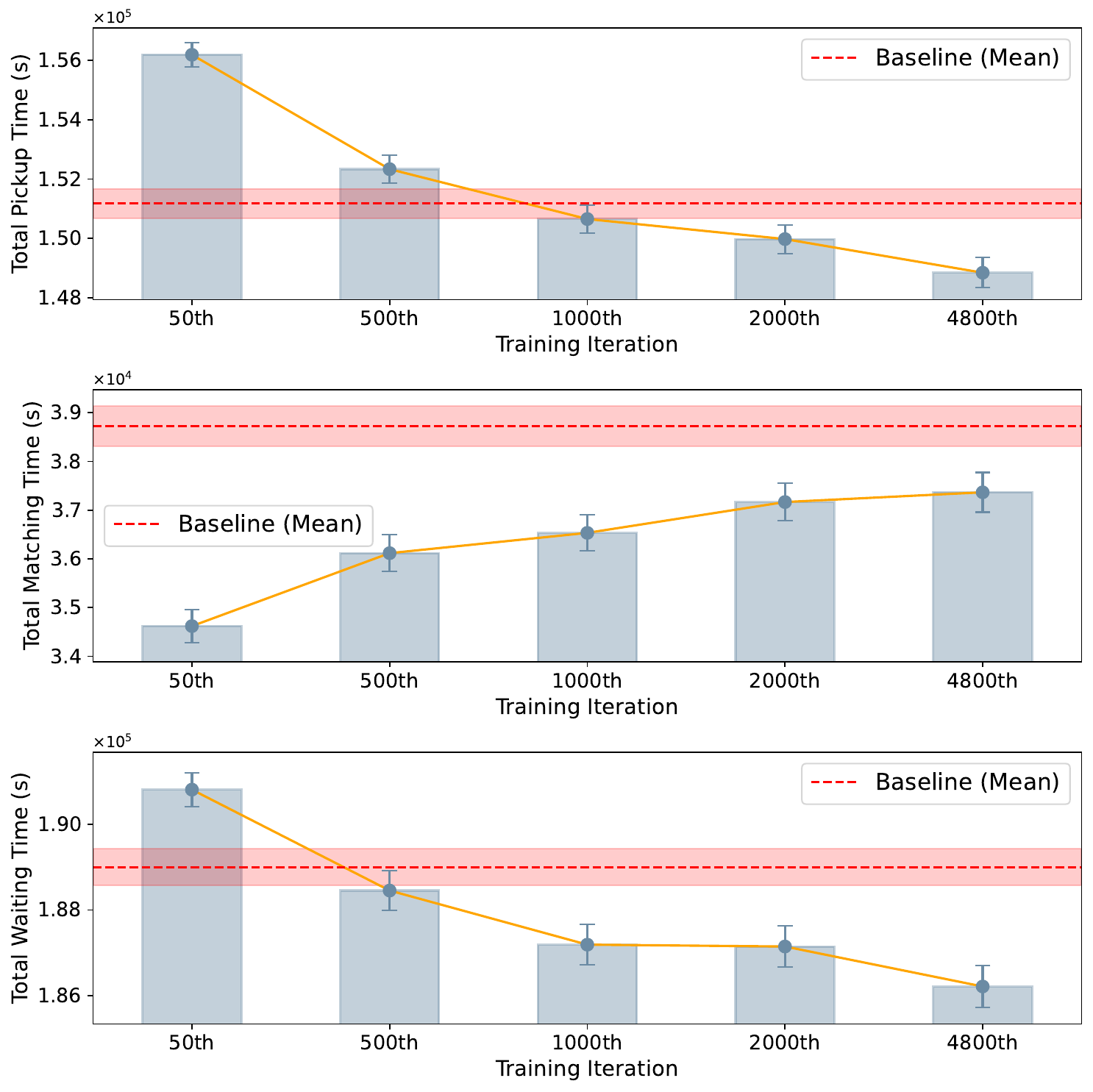} 
    \caption{Comparison of System Waiting Time Across Training Episodes and Baseline Strategy for Ride-Hailing Services} 
    \label{fig:/metrics_ride_hailing_system} 
\end{figure}

% \appendices

% % you can choose not to have a title for an appendix
% % if you want by leaving the argument blank
% \section{}

% \section*{Acknowledgment}


\ifCLASSOPTIONcaptionsoff
  \newpage
\fi



%\bibliographystyle{IEEEtran}
\bibliographystyle{IEEEtran} 
\bibliography{main} 
% \end{thebibliography}


\begin{IEEEbiography}{Yiman Bao}
\end{IEEEbiography}

\begin{IEEEbiographynophoto}{Jie Gao}
\end{IEEEbiographynophoto}

%\newpage

\begin{IEEEbiographynophoto}{Jinke He}
\end{IEEEbiographynophoto}


%\vfill


\end{document}